\documentclass{article} %
\PassOptionsToPackage{sort}{natbib}
\usepackage{iclr2027_conference,times}

\usepackage{amsmath,amsfonts,bm}

\def\eqref#1{equation~\ref{#1}}
\def\1{\bm{1}}

\DeclareMathAlphabet{\mathsfit}{\encodingdefault}{\sfdefault}{m}{sl}
\SetMathAlphabet{\mathsfit}{bold}{\encodingdefault}{\sfdefault}{bx}{n}

\usepackage{url}

\usepackage{graphicx}
\usepackage[dvipsnames]{xcolor}
\usepackage{subcaption}
\usepackage[font=small]{caption}
\definecolor{linkpurple}{HTML}{7069B5}
\usepackage[pagebackref=true,colorlinks,citecolor=brown]{hyperref}
\usepackage{xcolor}
\usepackage{amsmath}
\usepackage{amssymb}
\usepackage{algorithm}
\usepackage{algpseudocode}
\usepackage{wrapfig} 
\usepackage{multirow}
\usepackage{booktabs} 
\title{Scaling Transformer Energy-based Models
For High Quality Image Synthesis}

{\centering
\author{
  \textbf{Jungsoo Lee}\textsuperscript{1}\thanks{bus6479@mit.edu}
  \quad
  \textbf{Alexi Gladstone}\textsuperscript{2}
  \quad
  \textbf{Yilun Du}\textsuperscript{3}
  \\[4pt]
  \textsuperscript{1}MIT
  \quad
  \textsuperscript{2}UIUC
  \quad
  \textsuperscript{3}Harvard University 
  \\[6pt]
  \href{https://int-lee.github.io/scalable-ebm/}{%
 {\small\textcolor{magenta}{\textbf{Project page}}}}
  \qquad
  \href{https://github.com/int-lee/TEM}{%
   {\small\textcolor{magenta}{\textbf{Code}}}}
}
}

\newlength{\tilew}
\newlength{\bigw}
\newsavebox{\gridbox}

\iclrfinalcopy %
\begin{document}

\maketitle

\begin{abstract}
Energy-Based Models (EBMs) are a flexible class of generative models that decouple probabilistic modeling from the generation process. This decoupling directly enables compositional generation and inference-time scaling, where sample quality can be refined simply by increasing sampling steps. However, severe challenges in high-dimensional sampling and training instabilities with deep architectures have restricted EBMs to low-dimensional regimes. To resolve these bottlenecks, we introduce the Transformer Energy-Based Model (TEM), a scalable transformer energy function operating entirely within a latent space. On top of this backbone, we extract energies at multiple layers and sum them, composing complementary constraints from local texture to global semantics. Furthermore, two components make this high-quality at scale: Positive Representation Alignment (PRA), which supervises TEM's intermediate representations with a vision foundation model, and advanced sampling strategies that improve sample quality at inference. TEM-XL achieves an FID-50K of 3.82 on class-conditional ImageNet 256$\times$256 without classifier-free guidance, showing that a contrastive EBM can generate high-fidelity images. On ImageNet 512$\times$512, TEM achieves an FID-50K of 8.24, demonstrating that the approach scales to higher resolution. We further validate the core promises of EBMs: compositional generation, zero-shot inpainting, and out-of-distribution detection.
\end{abstract}

\section{Introduction}

Energy-Based Models (EBMs)~\citep{lecun2006tutorial, du2019implicit} parameterize probability distributions via an energy function that assigns lower energies to more probable samples and higher energies to less probable samples.
EBMs provide a flexible and mathematically elegant framework that fundamentally separates explicit generation from probabilistic modeling. Unlike Diffusion~\citep{ho2020denoising, sohl2015deep} and Flow~\citep{lipman2022flow} models, which are mathematically constrained to pre-defined generative trajectories governed by carefully designed timestep schedules, EBMs define the data distribution solely through an energy function, entirely decoupling the model from the generative process. This decoupling of the model from the generative process yields profound advantages. First, EBMs decouple learning from inference, allowing the sampling procedure to be chosen independently of model training. This enables flexible test-time compute scaling by increasing the MCMC steps, progressively refining the samples through a deeper exploration of the learned energy landscape. Second, this flexibility enables the composition of different distributions together~\citep{du2019implicit,du2023reduce}, which can enable zero-shot generation and generalization capabilities.

However, despite their theoretical advantages and compositional properties, EBMs have struggled with widespread adoption due to several critical unresolved challenges. First, EBMs face severe vulnerability to the curse of dimensionality: MCMC sampling suffers from poor mixing in high-dimensional spaces, leading to slow convergence and mode collapse. Second, EBMs scale poorly with model capacity. As network depth increases, conventional architectures exhibit severe training instability, making robust optimization exceedingly difficult~\citep{du2019implicit,du2020improved}. Consequently, existing EBM research has been largely limited to smaller-scale benchmarks, unable to demonstrate competitive performance on challenging, high-resolution datasets.

\begin{figure}[t]
\centering
\sbox{\gridbox}{%
\begin{minipage}[c]{0.62\textwidth}
\vspace{0pt}
\centering
\setlength{\tabcolsep}{1pt}
\renewcommand{\arraystretch}{0}
\setlength{\tilew}{0.155\linewidth}
\setlength{\bigw}{\dimexpr 2\tilew + 2\tabcolsep\relax}
\begin{tabular}{cccc}
\includegraphics[width=\bigw,height=\bigw]{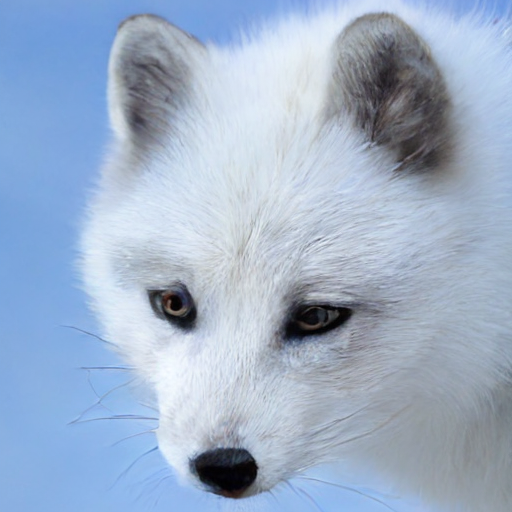} &
\includegraphics[width=\bigw,height=\bigw]{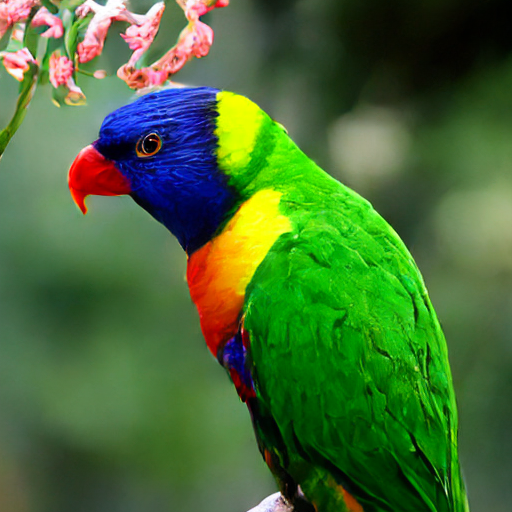} &
\begin{tabular}[b]{@{}c@{}}
  \includegraphics[width=\tilew,height=\tilew]{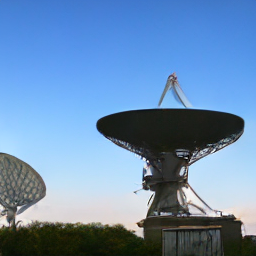} \\[2pt]
  \includegraphics[width=\tilew,height=\tilew]{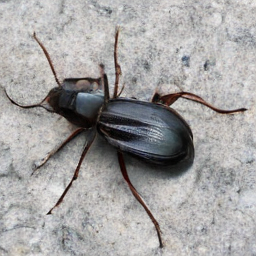}
\end{tabular} &
\begin{tabular}[b]{@{}c@{}}
  \includegraphics[width=\tilew,height=\tilew]{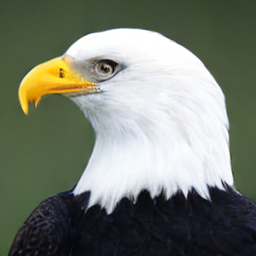} \\[2pt]
  \includegraphics[width=\tilew,height=\tilew]{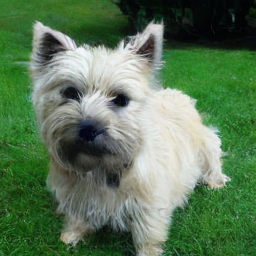}
\end{tabular}
\end{tabular}

\vspace{2pt}

\begin{tabular}{cccccc}
\includegraphics[width=\tilew,height=\tilew]{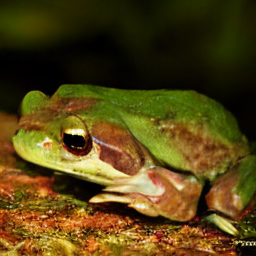} &
\includegraphics[width=\tilew,height=\tilew]{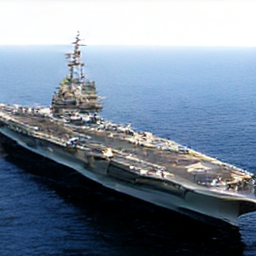} &
\includegraphics[width=\tilew,height=\tilew]{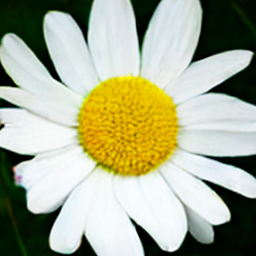} &
\includegraphics[width=\tilew,height=\tilew]{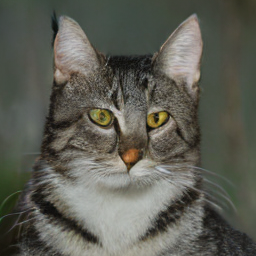} &
\includegraphics[width=\tilew,height=\tilew]{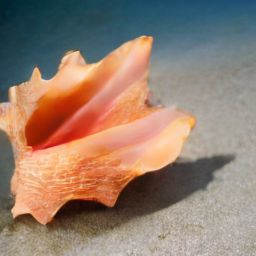} &
\includegraphics[width=\tilew,height=\tilew]{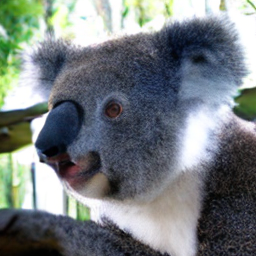}
\end{tabular}
\end{minipage}}
\usebox{\gridbox}%
\hspace{0.03\textwidth}%
\resizebox{!}{\ht\gridbox}{%
\begin{minipage}[c]{0.35\textwidth}
\vspace{0pt}
\centering
\setlength{\tabcolsep}{2pt}
\begin{tabular}{cc}
\includegraphics[width=0.46\linewidth]{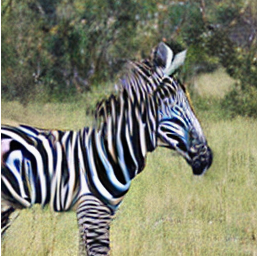} &
\includegraphics[width=0.46\linewidth]{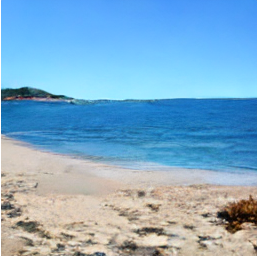} \\[1pt]
\small \textit{``zebra''} & \small \textit{``beach''}
\end{tabular}

\vspace{1pt}

{\small $\boldsymbol{\Downarrow}$}

\vspace{1pt}

\includegraphics[width=0.46\linewidth]{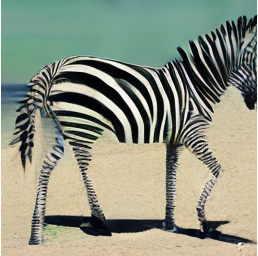}\\[1pt]
\footnotesize \textit{composed result}
\end{minipage}}

\caption{\textbf{High-fidelity ImageNet 256$\times$256, 512$\times$512 generation and compositional synthesis by TEM-XL/2.}
Left: uncurated class-conditional samples from 80 Langevin steps. The two large images are 512$\times$512 samples and the remaining tiles are 256$\times$256.
Right: compositional generation by summing the energy functions of two independent classes, producing a coherent scene that satisfies both concepts without retraining.}
\label{fig:Teaser}
\vspace{-10pt}
\end{figure}

In this work, we address these fundamental limitations and demonstrate that EBMs
trained contrastively, with negatives drawn from the model's own MCMC
sampler, can generate high-quality $256 \times 256$
ImageNet~\citep{deng2009imagenet} samples through persistent sampling initialized solely from Gaussian noise,
achieving performance competitive with strong generative models (Table~\ref{tab:all_results}). We further validate the core theoretical promises of EBMs by demonstrating compositional generalization as well as flexible inference-time control. Our work addresses the long-standing scalability bottleneck of contrastive EBMs, reestablishing them as a viable contender for high-fidelity image generation.

Our key contributions are: 
\textbf{(1)} A scalable Vision Transformer-based EBM architecture with multiple energy heads that enables stable training at high resolution.
\textbf{(2)} A comprehensive training framework encompassing latent space design, representation alignment, and advanced sampling strategies that together achieve state-of-the-art quality images for contrastive energy-based models.
\textbf{(3)} Empirical validation of the theoretical advantages of energy-based
modeling at high resolution: compositional generation, zero-shot inpainting, and
out-of-distribution detection from a single scalar energy.

\section{Background}

\subsection{Energy-Based Models}
Energy-Based Models (EBMs) define a probability distribution implicitly over an input domain using a scalar energy function, expressed as $p_{\theta}(\mathbf{x}) = \frac{\exp(-E_{\theta}(\mathbf{x}))}{Z(\theta)}$, where $E_{\theta}(\mathbf{x})$ is the parameterized energy function. 
In practice, $E_{\theta}(\mathbf{x})$ is parameterized by a neural network $E_{\theta} : \mathcal{X} \to \mathbb{R}$, yielding an unnormalized density model. Maximum likelihood training then shapes the energy landscape such that low-energy regions correspond to high-density regions of the data distribution $p_{\text{data}}(\mathbf{x})$, ensuring that $p_{\theta}(\mathbf{x})$ approximates $p_{\text{data}}(\mathbf{x})$. However, $Z(\theta) = \int_{\mathbf{x}} \exp(-E_{\theta}(\mathbf{x})) d\mathbf{x}$ is intractable, which makes exact maximum likelihood training of EBMs computationally infeasible. On the other hand, the gradient of log-likelihood remains tractable~\citep{du2019implicit,hinton2006unsupervised}:
\begin{equation}
\nabla_{\theta} \mathbb{E}_{x \sim p_{\text{data}}(x)} \big[ -\log p_{\theta}(x) \big]
=
\mathbb{E}_{x \sim p_{\text{data}}(x)} \big[  \nabla_{\theta} E_{\theta}(x) \big]
-
\mathbb{E}_{x \sim p_{\theta}(x)} \big[  \nabla_{\theta} E_{\theta}(x) \big]
\label{eq:mle-grad}
\end{equation}
The first term is estimated directly from samples from the data. The second requires expectations under $p_\theta(x)$, which is intractable to
sample exactly. Running Markov Chain Monte Carlo (MCMC) to convergence is
required, implying infinitely long chains with infinitesimal steps.

\subsection{Short-Run Langevin MCMC}

 We therefore rely on short-run MCMC to obtain approximate samples. In particular, we start from a fixed initial distribution $p_0(\mathbf{z})$ and perform a finite number of $T$ Langevin updates with step size $\eta$:
\begin{equation}
 \quad
\mathbf{z}_{t+1} = \mathbf{z}_t + \eta \nabla_{\mathbf{z}} \log \pi(\mathbf{z}_t) + \sqrt{2\eta}\,\boldsymbol{\epsilon},
\quad t = 0, \ldots, T-1,
\label{eq:langevin}
\end{equation}
where $\boldsymbol{\epsilon} \sim \mathcal{N}(0, \mathbf{I})$~\citep{du2019implicit, nijkamp2019learning}. Let $\tilde{\pi}(\mathbf{z})$ denote the distribution of the final state $\mathbf{z}_T$. Since initialization $p_0$, step size $\eta$, and number of steps $T$ are fixed, $\tilde{\pi}$ is well-defined. In practice, we use a finite number of steps and tune the step size and noise scale independently, yielding an efficient but biased sampler.

\section{Scalable Transformer Energy-Based Models}
\begin{figure}[t]
    \centering
    \begin{minipage}{0.96\linewidth}
        \centering
        \begin{subfigure}[t]{0.48\linewidth}
            \centering
            \includegraphics[width=\linewidth]{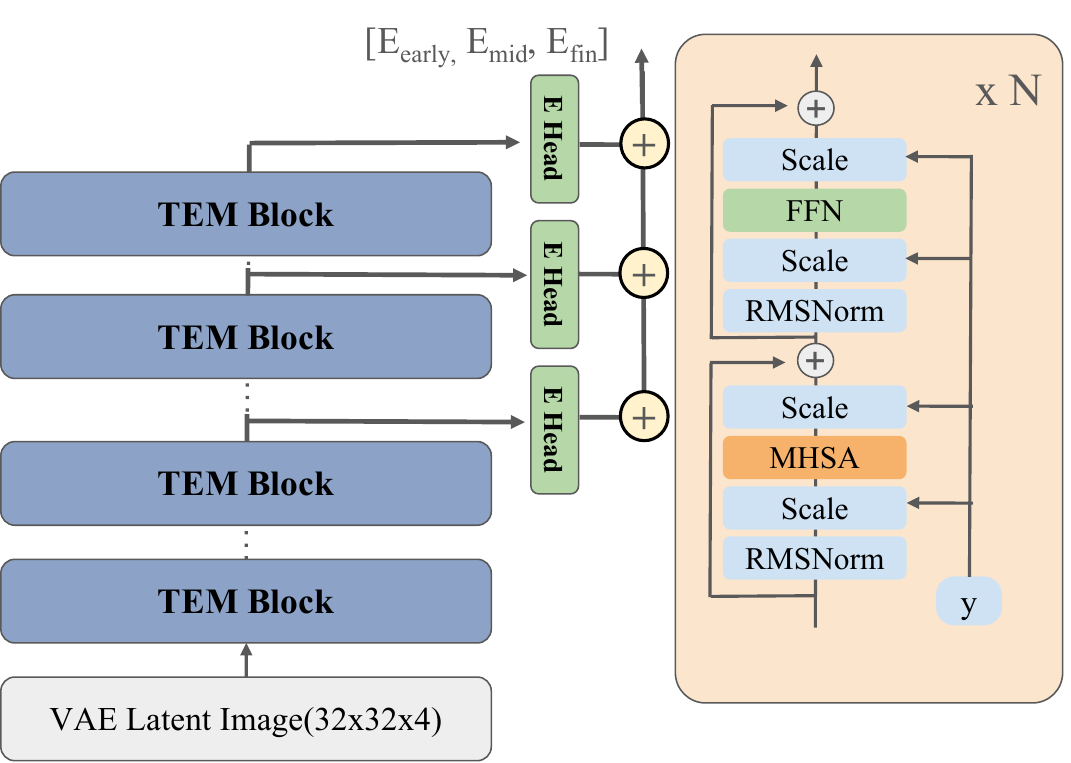}
     
        \end{subfigure}
        \hfill
        \begin{subfigure}[t]{0.48\linewidth}
            \centering
            \includegraphics[width=\linewidth]{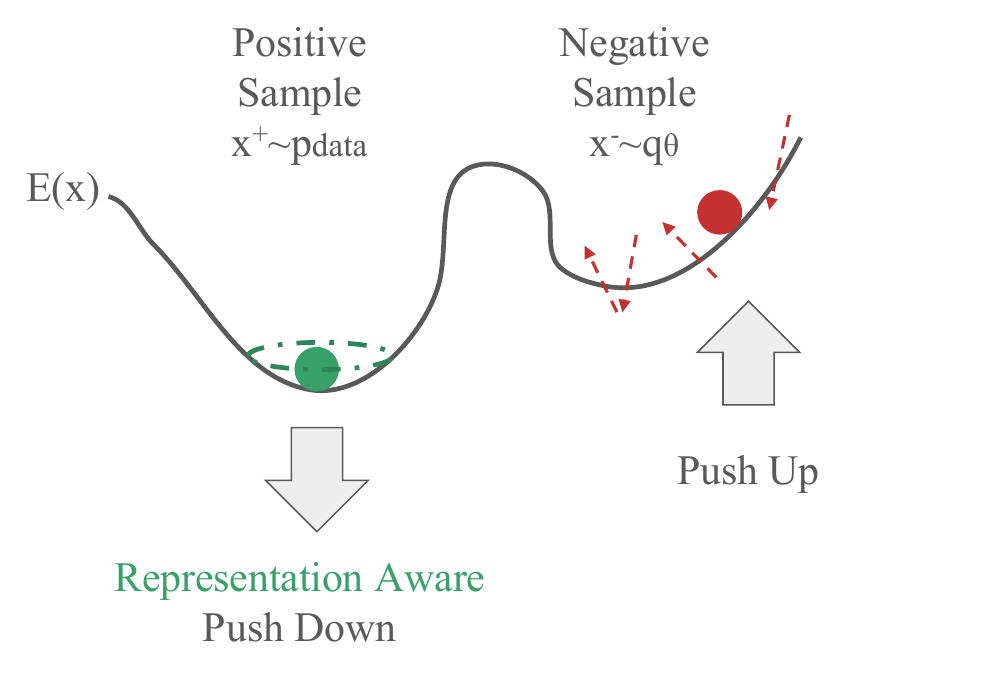}
       
        \end{subfigure}
    \end{minipage}
    \caption{\textbf{Proposed Architecture and Training Strategy.} \textit{Left}: the Transformer Energy-Based Model processes VAE-encoded latent tokens through a stack of TEM blocks. Energy heads at multiple intermediate layers produce scalar energies whose sum defines the total energy. \textit{Right}: contrastive training pushes down the energy of data samples~$x^+$ and pushes up the energy of MCMC negatives~$x^-$, while Positive Representation Alignment (PRA) aligns the model's internal features with a frozen VFM.}
    \label{fig:ArchandAlgo}
    \vspace{-15pt}
\end{figure}

\subsection{Transformer Energy-Based Models Architecture}
\textbf{TEM Architecture.} 
We parameterize the energy function $E_\theta(x,c)$ with a Vision Transformer (ViT)~\citep{dosovitskiy2020image} backbone adapted for large-scale energy-based modeling. 
Given an input $x$, we tokenize it into spatial patch embeddings, which are processed through a stack of transformer blocks. The final patch tokens are aggregated via global average pooling and passed through a small MLP head to produce a scalar energy $E_\theta(x,c)$.

Our architecture adopts several modern design choices for stability and expressivity. We encode spatial structure with rotary positional embeddings~\citep{su2024roformer},  and apply query-key normalization~\citep{henry2020query} to stabilize attention logits at scale. Feed-forward layers use SwiGLU activations~\citep{shazeer2020glu}, and the conditioning signal is projected into feature-wise affine parameters in an AdaLN-style formulation~\citep{peebles2023scalable} built on the same RMSNorm variant.

Importantly, this design demonstrates stable and effective energy-based modeling at scale without requiring EBM specific components such as spectral normalization commonly used in prior energy-based models~\citep{du2019implicit}.

\textbf{Multi-layer Energy Extraction.} Energy-based models are naturally suited to composition, since combining distributions amounts to addition in energy space. \citet{du2020improved} demonstrate this by composing several separately trained models, each capturing a distinct concept. Composition of this kind has so far required multiple models.

We show that it can instead be obtained within one, extracting energies through multiple heads attached to representations at different depths of a single Transformer. Its layers encode progressively more abstract features, from low-level structure in early layers to high-level semantics in deeper ones~\citep{raghu2021vision}. For $L$ depth layer with $k$ heads, we attach them at the blocks nearest to ($L/k$, $2L/k$, \ldots, $L$). The resulting heads form a hierarchy of energy terms, each enforcing the structure visible at its depth. We train on the sum of these energies, allowing each layer to contribute at
its own level of abstraction. This not only yields a better energy function
but also decomposes the generative process across layers, opening the way to
understanding each layer's role and further applications.

\subsection{Training Transformer Energy-Based Models}
\label{sec:training_ebm}

\textbf{Latent Space.} Training $E_\theta$ requires sampling negatives by MCMC at every step, which is expensive in high dimensions and mixes poorly. %
Therefore, we operate entirely in the latent space induced by the autoencoder of Stable Diffusion~\citep{rombach2022high}, which is common in the literature~\citep{peebles2023scalable}. Given an input image $x \in \mathbb{R}^{256 \times 256 \times 3}$, we encode to latent space $z \in \mathbb{R}^{32 \times 32 \times 4}$ where the encoder predicts the mean $\mu_{enc}(x)$ and variance $\sigma_{enc}^2(x)$. We sample latent variables as
\begin{equation}
z \sim \mathcal{N}(\mu_{enc}(x), \sigma_{enc}^2(x))
\end{equation}
which corresponds to drawing from the learned posterior distribution of the autoencoder. In this latent space, we define an energy-based model $E_\theta(z, c)$ and perform learning and sampling directly on $z$. The decoder $D(\cdot)$ maps latent samples back to the image space via $x \approx D(z)$.

\textbf{Contrastive Energy Objective.}
We train the energy-based model using a contrastive objective that encourages lower energy for data samples than for model samples.
\begin{equation}
\mathcal{L}_{\text{contrast}} =  E(z^+)  - E(q_{\theta}(z_0)) +\lambda \left\{ E(z^+)^2 + E(q_{\theta}(z_0))^2\right\}
\label{eq:contrast}
\end{equation}
Here, $z^+$ denotes an encoded positive sample drawn from the dataset, and $q_{\theta}(z_0)$ denotes a negative sample produced by sampler.  $z_0$ is from pure noise or replay buffer. Since the likelihood gradient depends only on energy differences, the magnitudes of energies are relative and drift during training. We add an L2 regularization term penalizing the magnitude of the energy to resolve this gauge freedom and bound the energy magnitude following~\citet{du2019implicit}.

\textbf{Sampler Matching Loss.} Because the sampler runs for a finite number of steps, its output distribution $q_\theta$ may not exactly match the model distribution $p_\theta$ from which the negative phase should draw. Following~\citet{du2020improved}, we introduce a regularizer motivated by $\mathrm{KL}(q_\theta \,\|\, p_\theta)$:
\begin{align}
\mathcal{L}_{\text{sampler-match}}
&=
\underbrace{
\mathbb{E}_{q_\theta(\hat{\mathbf{z}}^-)}\!\left[
E_{\mathrm{sg}(\theta)}(\hat{\mathbf{z}}^-)
\right]
}_{\mathcal{L}_{\mathrm{opt}}}
+
\underbrace{
\mathbb{E}_{q_\theta(\hat{\mathbf{z}}^-)}\!\left[
\log q_\theta(\hat{\mathbf{z}}^-)
\right]
}_{\mathcal{L}_{\mathrm{ent}}},
\label{eq:correction}
\end{align}
where $\hat{\mathbf{z}}^-$ is a sample generated by sampler and the stop-gradient $\mathrm{sg}(\cdot)$ freezes the energy parameters, so gradients reach $\theta$ only through the sampler. The first term drives the sampler toward low-energy regions. The second is an entropy term, estimated with a nearest-neighbor statistic~\citep{beirlant1997nonparametric}, that keeps its outputs diverse. Together, these terms encourage low-energy, diverse sampler outputs. Appendix~\ref{sec:derivationkl},~\ref{sec:trainingdetail} gives the details and reports training settings.

\textbf{Representation Supervision.} Contrastive EBM training provides only a scalar supervisory signal per sample, one energy value, which makes representation learning challenging at high resolution. A natural remedy is to supply a dense signal to the representation itself, as in diffusion models, where aligning generative features with those of a pretrained vision foundation model (VFM) substantially improves sample quality~\citep{yu2024representation, wang2025diffuse}. We study how such alignment should be formulated for energy-based models and introduce \emph{Positive Representation Alignment} (PRA), an objective designed for the contrastive setting.

Let $\bar{h}_{\mathrm{VFM}}(x^+) \in \mathbb{R}^{N \times D}$ denote the spatially normalized patch-wise output of the VFM, where $N$ is the number of patches and $D$ is the embedding dimension.
Let $h_\theta(z^+)$ denote the internal representation of the EBM and $P_\phi(\cdot)$ a learnable projection network.
PRA maximizes the patch-wise cosine similarity between the projected EBM features and the VFM targets:
\begin{equation}
\mathcal{L}_{\mathrm{PRA}}(\theta, P_\phi)
:=
-\frac{1}{N}
\sum_{n=1}^{N}
\mathrm{sim}\!\left(
\bar{h}_{\mathrm{VFM}}(x^+)_{[n]}, \,
P_\phi\big(h_\theta(z^+)\big)_{[n]}
\right).
\label{eq:pra}
\end{equation}
We apply this alignment loss only to positive (data) samples, and never to
MCMC negatives. Two complementary intuitions motivate this choice. From a contrastive
learning view, alignment acts as a shaping term on the low-energy side of the
landscape, while the objective is simultaneously driving negatives toward high
energy. Aligning negatives would therefore pull them back toward the data
manifold in feature space, working directly against that push. From a manifold
view, the VFM produces meaningful features only for inputs near the natural-data
manifold. MCMC negatives are typically off-manifold, and this is most pronounced
early in training and at intermediate sampling steps, so their embeddings
provide no reliable semantic target to align against.

This positive-only design also clarifies how PRA is both similar and different to representation
alignment in diffusion models, where the distinction lies in what each network
receives as input. That line of work aligns the features of a noisy input $x_t$
with the VFM features of its clean counterpart $x_0$; the clean target is
dynamic and guides the denoising trajectory toward the data manifold. In our
setting both the EBM and the VFM receive the same clean input $x^+$, so PRA does
not guide any sampling trajectory. Instead it shapes the geometry of the low-energy region of the
landscape directly, PRA encourages semantically structured representations of positive samples.

\textbf{Overall Objective.}
Combining the contrastive energy objective (Eq.~\ref{eq:contrast}), the sampler matching term (Eq.~\ref{eq:correction}), and the representation alignment loss (Eq.~\ref{eq:pra}), the full training objective is:
\begin{equation}
\mathcal{L}_{\text{total}} = 
\mathcal{L}_{\text{contrast}} 
+ \lambda_1\mathcal{L}_{\text{sampler-match}} 
+ \lambda_2\mathcal{L}_{\text{PRA}},
\label{eq:overall}
\end{equation}
where $\lambda_1$ and $\lambda_2$ are coefficients for each loss.
Each term plays a distinct role: $\mathcal{L}_{\text{contrast}}$ shapes the energy landscape by pushing data samples to low energy and sampled negatives to high energy, $\mathcal{L}_{\text{PRA}}$ injects semantic structure into the model's internal representations, and $\mathcal{L}_{\text{sampler-match}}$ encourages low-energy, diverse sampler outputs.
The full training algorithm is provided in Appendix~\ref{training algorithm}.

\subsection{Inference Time Scaling}

At inference, we adjust the sampling step size while keeping
the injected noise scale fixed:
\begin{equation}
x_{t+1} = x_t - \eta \nabla_x E_\theta(x_t)
+ \sigma\epsilon,
\qquad \epsilon \sim \mathcal{N}(0, I).
\label{eq:langevin_temp}
\end{equation}
 $\eta$ controls
the balance between energy-driven refinement and stochastic
exploration. It provides control over sample quality, without retraining or additional unconditional
network evaluations.

Increasing the step size $\eta$ amplifies the deterministic
energy-gradient update relative to the fixed noise.
Similarly, extending sampling beyond the training horizon
allows further refinement of intermediate samples.
Empirically, both adjustments can sharpen individual samples
and produce more class-typical images at the cost of mode
coverage.

\begin{table*}[t]
\centering
\small
\caption{\textbf{Class-conditional generation on ImageNet-256$\times$256 and 512$\times$512 (FID-50K).}
Params refer to the 256$\times$256 configuration unless a resolution-specific model is used. Baseline values are taken as reported in the original papers; ``--'' indicates that the corresponding setting is or was not reported.}%
\vspace{-0.3cm}
\renewcommand{\arraystretch}{1.2}
\setlength{\tabcolsep}{5pt}
\begin{tabular}{lcccccccc}
\toprule
& & \multicolumn{3}{c}{ImageNet 256$\times$256} & \multicolumn{3}{c}{ImageNet 512$\times$512} \\
\cmidrule(lr){3-5} \cmidrule(lr){6-8}
Method & Params & NFE & FID$\downarrow$ & FID$\downarrow$ (CFG) & NFE & FID$\downarrow$ & FID$\downarrow$ (CFG) \\
\midrule
\multicolumn{8}{l}{\textit{\textbf{GAN}}} \\
BigGAN-deep & 112M & 1 & 6.95 & -- & 1 & 8.43 & -- \\
GigaGAN & 569M & 1 & 3.45 & -- & -- & -- & -- \\
\midrule
\multicolumn{8}{l}{\textit{\textbf{Autoregressive / Masked}}} \\
MaskGIT & 227M & 8 & 6.18 & -- & 12 & 7.32 & -- \\
VAR-d30 & 2B & 10 & -- & 1.92 & -- & -- & -- \\
\midrule
\multicolumn{8}{l}{\textit{\textbf{Diffusion/Flow}}} \\
ADM & 554M & 250 & 10.94 & -- & 250 & 23.24 & -- \\

LDM-4 & 400M & 250 & 10.56 & 3.60 & -- & -- & -- \\
DiT-XL/2 & 675M & 250 & 9.62 & 2.27 & 250 & 12.03 & 3.04 \\

SiT-XL/2 & 675M & 250 & 8.3 & 2.06 & 250 & -- & 2.62 \\
SiT-XL/2+REPA & 675M & 250 & 5.9 & 1.42 & 250 & -- & 2.08 \\
\midrule
\multicolumn{8}{l}{\textit{\textbf{EBM}}} \\
TEM-L & 415M & 60 & 4.74 & -- & -- & -- & -- \\
TEM-XL & 609M & 60 & 3.82 & -- & 60 & 8.24 & -- \\
\bottomrule
\end{tabular}
\vspace{-0.5cm}
\label{tab:all_results}
\end{table*}

\begin{figure}[t]
    \centering
    \begin{minipage}{\linewidth}
        \centering
        \begin{subfigure}[t]{0.32\linewidth}
            \centering
            \includegraphics[width=\linewidth]{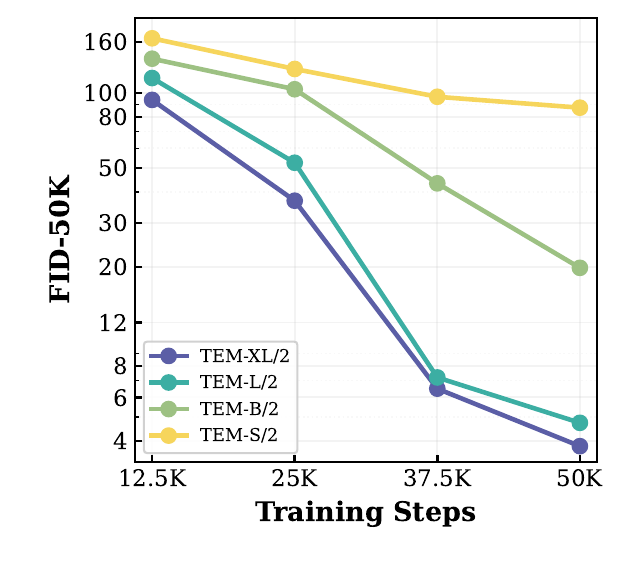}
        \end{subfigure}
        \hfill
        \begin{subfigure}[t]{0.32\linewidth}
            \centering
            \includegraphics[width=\linewidth]{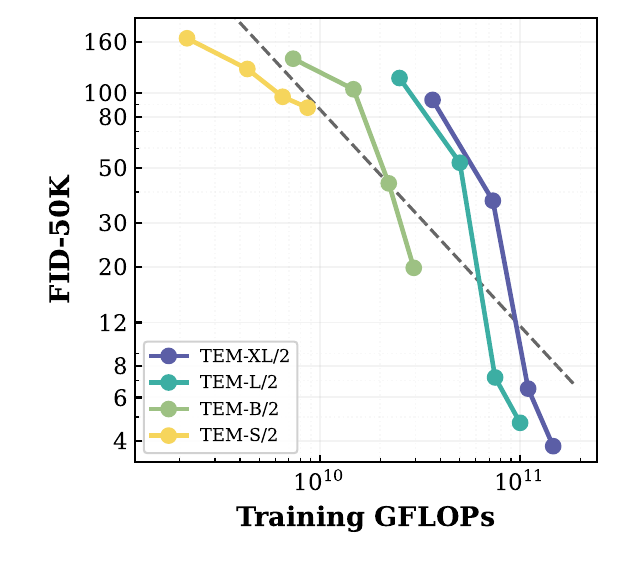}
        \end{subfigure}
        \hfill
        \begin{subfigure}[t]{0.32\linewidth}
            \centering
            \includegraphics[width=\linewidth]{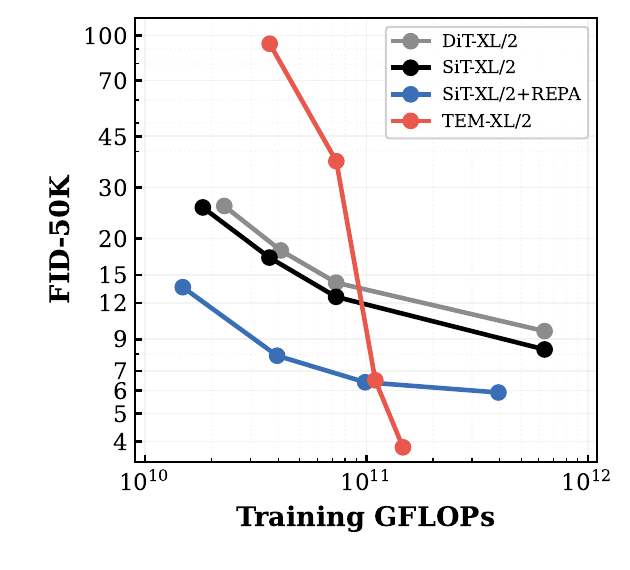}
        \end{subfigure}
    \end{minipage}
    \vspace{-0.2cm}
    \caption{\textbf{Scaling behavior of TEM on class-conditional ImageNet 256$\times$256.}
    \textit{Left}: FID-50K as a function of training steps across model sizes. Larger models consistently achieve lower FID at every checkpoint; 40 sampling steps are used for sampling during training.
    \textit{Middle}: FID-50K as a function of cumulative training GFLOPs. All four model sizes follow a consistent compute--quality trend (dashed line).
    \textit{Right}: FID-50K versus training GFLOPs for TEM-XL compared against DiT-XL, SiT-XL and SiT-XL+REPA showing that TEM tracks, and eventually achieves lower FID than unguided diffusion baselines with less training compute.}
    \label{fig:diff_vs_ebm}
\vspace{-0.2cm}
\end{figure}

\section{Experiments}
\subsection{Main Results}

Unless otherwise noted, all TEM experiments use batch size 256, learning rate
$2\times10^{-4}$, and 50K iterations, training with 40 steps sampling and inference with 60 steps sampling. Main benchmark numbers for TEM are obtained with
inference-time scaling but without classifier-free guidance. All conditions
within each ablation are controlled. Full training details are provided in Appendix~\ref{sec:trainingdetail}.

\begin{wrapfigure}[17]{r}{0.40\textwidth}
    \centering
    \vspace{-20pt}
    \includegraphics[width=0.40\textwidth]{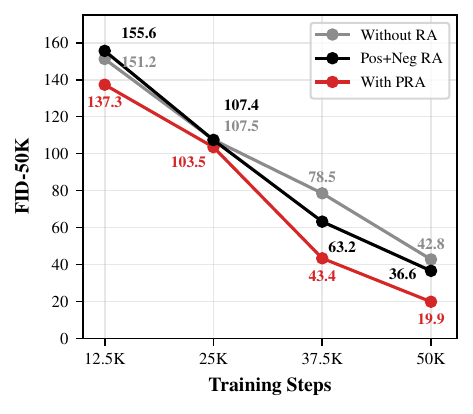}
    \vspace{-20pt}
    \caption{\textbf{Effect of PRA on training.} FID-50K across training steps for
TEM-B with no representation alignment, Pos+Neg RA, and PRA. PRA consistently
outperforms both, reaching 19.9 FID at 50K steps versus 42.8 without alignment
and 36.6 with Pos+Neg RA. This supports aligning only positive samples in our
method. }
    \label{fig:fid_comparison}
    \vspace{20pt}
\end{wrapfigure}

Our TEM-XL model achieves a guidance-free FID-50K of 3.82 on ImageNet
256$\times$256 and 8.24 on 512$\times$512 (Table~\ref{tab:all_results}).
To our knowledge, these are the best reported large-scale results for a contrastively trained EBM using its own MCMC negatives. Evaluation uses persistent sampling initialized from Gaussian noise and outperforms unguided diffusion transformers of comparable size. This brings contrastive EBMs into the performance range of
modern large-scale generative models. Although EBMs are often considered expensive, TEM's training and inference costs are comparable to, or even lower than, those of diffusion transformers of similar size (Appendix~\ref{app:cost}).
As shown in Figure~\ref{fig:diff_vs_ebm}, TEM scales with model size in both training steps and training compute (left, middle), and scales more steeply than other baselines, overtaking them at the highest budget (right).

\subsection{Ablations}
\begin{figure*}[t]
    \centering
    \sbox0{%
    \begin{minipage}[c]{0.30\linewidth}
        \centering
        \begin{subfigure}[b]{0.47\linewidth}
            \centering
            \includegraphics[width=\linewidth]{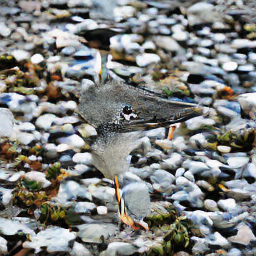}
            \vspace{-5mm}
            \caption{Early}
        \end{subfigure}
        \hfill
        \begin{subfigure}[b]{0.47\linewidth}
            \centering
            \includegraphics[width=\linewidth]{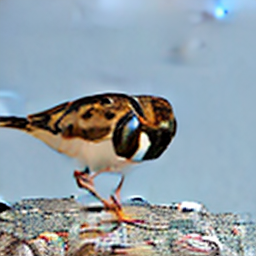}
            \vspace{-5mm}
            \caption{Middle}
        \end{subfigure}
        \vspace{1mm}
        \begin{subfigure}[b]{0.47\linewidth}
            \centering
            \includegraphics[width=\linewidth]{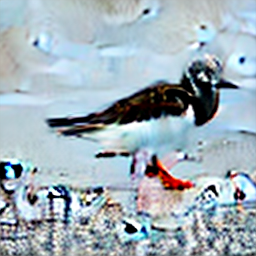}
            \vspace{-5mm}
            \caption{Final}
        \end{subfigure}
        \hfill
        \begin{subfigure}[b]{0.47\linewidth}
            \centering
            \includegraphics[width=\linewidth]{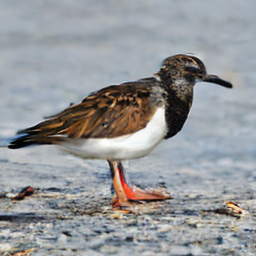}
            \vspace{-5mm}
            \caption{Full}
        \end{subfigure}
    \end{minipage}}
    \usebox0
    \hspace{-1mm}
    \begin{minipage}[c]{0.36\linewidth}
        \centering
        \vspace{-1mm}
        \includegraphics[width=\linewidth,
                         height=\dimexpr1.02\ht0+1.02\dp0\relax]{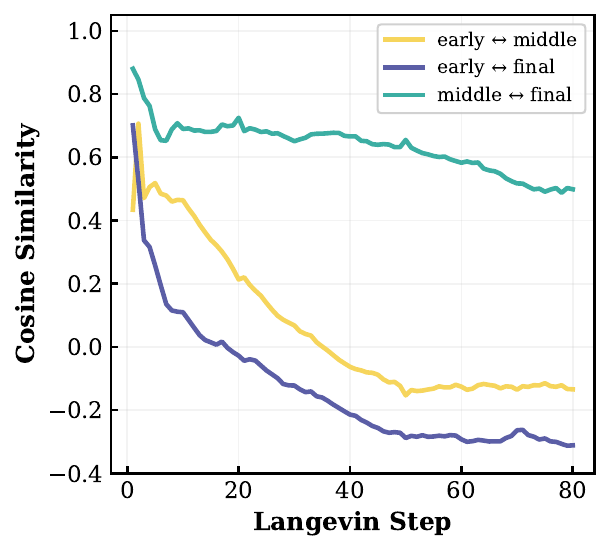}
    \end{minipage}
    \hspace{1mm}
    \begin{minipage}[c]{0.28\linewidth}
    \small
    \renewcommand{\arraystretch}{1.44}
    \vspace{-3mm}
    \begin{tabular*}{\linewidth}{@{\extracolsep{\fill}}lc@{}}
        \toprule
        \textbf{Variant} & \textbf{FID} $\downarrow$ \\
        \midrule
        \multicolumn{2}{@{}l}{\textit{Loss terms} (3 heads)} \\
         $\,\mathcal{L}_{\mathrm{contrast}}$ only & 121.11 \\
        $+\,\mathcal{L}_{\mathrm{opt}}$ & 46.30 \\
        $+\,\mathcal{L}_{\mathrm{ent}}$ & 42.76 \\
        $+\,\mathcal{L}_{\mathrm{PRA}}$ & \textbf{19.86} \\
        \midrule
        One head & 20.95 \\
        Two heads & 20.50 \\
        \bottomrule
    \end{tabular*}
\end{minipage}
    \vspace{-10pt}
    \caption{\textbf{Ablations on Multi-energy Head and Loss Terms.}
    \textit{Left}: generated images from different energy heads, where (a), (b), (c)
    decompose the energy into early, middle, and final heads respectively, and (d)
    uses the full energy head.
    \textit{Middle}: cosine similarity between energy head gradients across Langevin steps.
    \textit{Right}: FID-50K at 50K steps for TEM-B. The upper block adds each
loss term cumulatively with three heads; the lower block varies the number of
energy heads under the full loss, against the three-head configuration above.}
    \label{fig:multi-head}
    \vspace{-10pt}
\end{figure*}

\textbf{Analysis of Multi-Head Energy Composition.}
We validate the use of multiple energy functions through three analyses (Figure~\ref{fig:multi-head}). The three-head model reaches a lower FID than a single- or double-head baseline at 50K steps (right). Fitting three energy functions whose sum must jointly explain the data yields a better energy landscape.

\begin{figure}[t]
\centering
\begin{minipage}{1.0\textwidth}
\centering
\setlength{\tabcolsep}{2pt} %
\footnotesize
\begin{tabular}{ccccccc}
\shortstack{\includegraphics[width=0.13\linewidth]{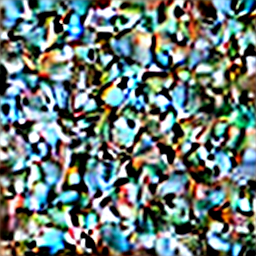}\\ \phantom{$T=0$}} &
\shortstack{\includegraphics[width=0.13\linewidth]{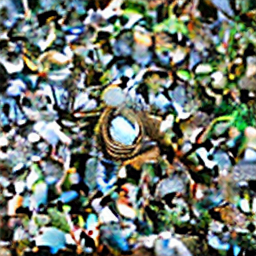}\\ \phantom{$T=20$}} &
\shortstack{\includegraphics[width=0.13\linewidth]{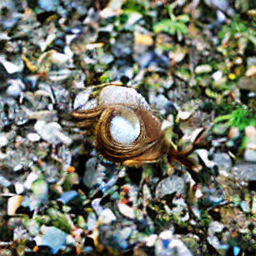}\\ \phantom{$T=40$}} &
\shortstack{\includegraphics[width=0.13\linewidth]{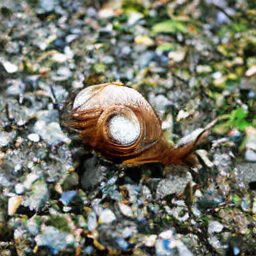}\\ \phantom{$T=60$}} &
\shortstack{\includegraphics[width=0.13\linewidth]{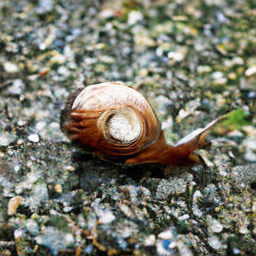}\\ \phantom{$T=80$}} &
\shortstack{\includegraphics[width=0.13\linewidth]{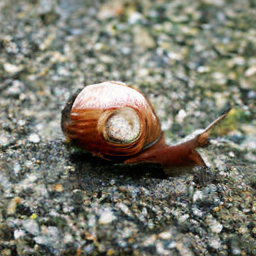}\\ \phantom{$T=100$}} &
\shortstack{\includegraphics[width=0.13\linewidth]{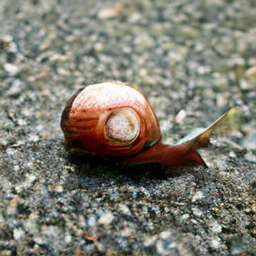}\\ \phantom{$T=120$}} \\[-5pt]
\shortstack{\includegraphics[width=0.13\linewidth]{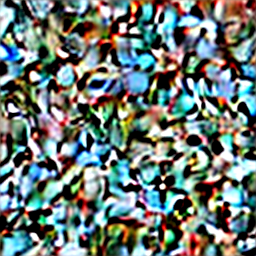}\\ $T=0$} &
\shortstack{\includegraphics[width=0.13\linewidth]{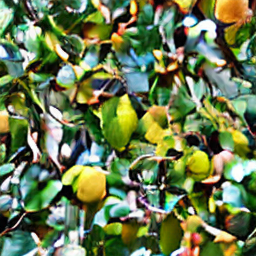}\\ $T=20$} &
\shortstack{\includegraphics[width=0.13\linewidth]{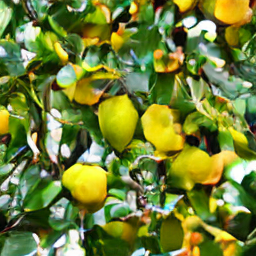}\\ $T=40$} &
\shortstack{\includegraphics[width=0.13\linewidth]{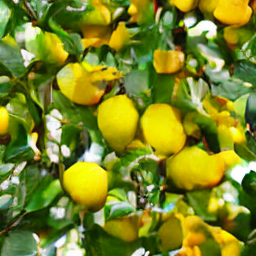}\\ $T=60$} &
\shortstack{\includegraphics[width=0.13\linewidth]{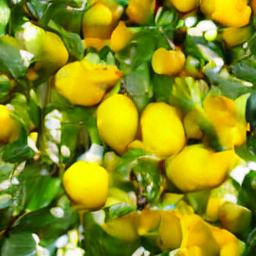}\\ $T=80$} &
\shortstack{\includegraphics[width=0.13\linewidth]{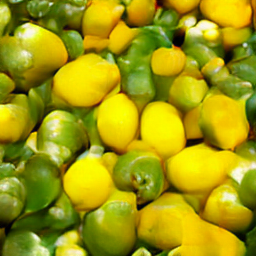}\\ $T=100$} &
\shortstack{\includegraphics[width=0.13\linewidth]{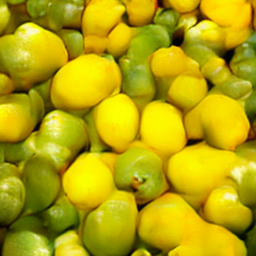}\\ $T=120$} \\
\end{tabular}
\captionof{figure}{\textbf{Progressive sample refinement via extended sampling steps.}
Two generation trajectories (top: snail, bottom: lemon) visualized from $T=0$ (pure noise) to $T=120$.
At the training horizon ($T=40$), coarse structure is established but residual noise remains. Extending beyond the training horizon produces increasingly sharp, class-typical images, demonstrating inference-time scaling without retraining.}
\label{fig:langevin_steps}
\end{minipage}
    \centering

    \begin{minipage}{0.44\linewidth}
        \centering
        
\begin{algorithm}[H]
\caption{Sampling Algorithm}
\label{alg:langevin}

\begin{algorithmic}[1]

\If{$|\mathcal{B}| =  N$} \Comment{$\mathcal{B} \gets \emptyset$ at beginning}
    \State Sample $x_0 \sim \mathcal{B}$
\Else
    \State Sample $x_0 \sim \mathcal{N}(0, I)$
\EndIf

\For{$t = 1$ to $T$}
    \State $x_t \leftarrow x_{t-1}
    - \eta \nabla_x E_\theta(x_{t-1})
    + \sigma \epsilon$
    \If{$t=40$}
    \State Add $x_{40}$ to replay buffer $\mathcal{B}$
    \EndIf

\EndFor

\State \Return $x_T$

\end{algorithmic}
\end{algorithm}
        \label{fig:left}
    \end{minipage}
    \hfill
    \begin{minipage}{0.54\linewidth}
        \centering
        \includegraphics[width=\linewidth]{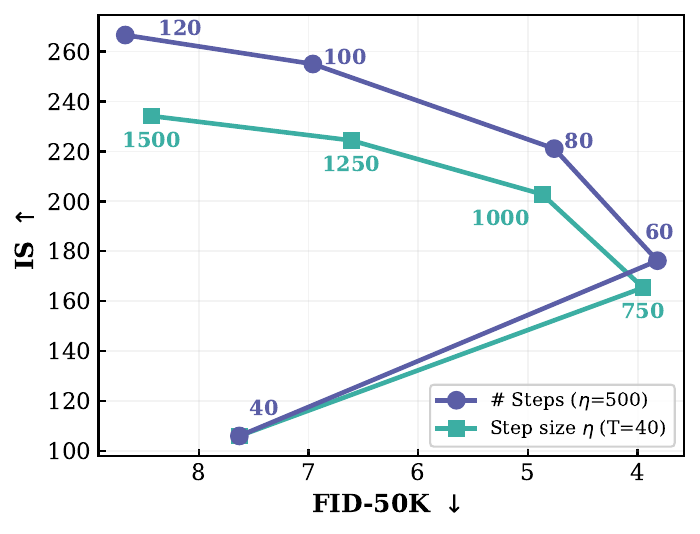}

        \label{fig:right}
    \end{minipage}

    \caption{\textbf{Inference-time scaling pseudocode and quality--diversity trade-off.}
    \textit{Left:} MCMC sampling procedure of TEM. Chains can run beyond 40 steps, but only the step-40 state is inserted into the replay buffer, so the buffer stays matched to the training dynamics.
\textit{Right:} FID-50K vs.\ Inception Score for TEM-XL, increasing the number of sampling steps (blue, $\eta{=}500$) or the step size (green, $T{=}40$) from the training configuration ($T{=}40$, $\eta{=}500$).  Moderately increasing the number of steps improves both metrics (FID 3.82 at 60 steps). Beyond that, IS keeps rising while FID degrades as samples concentrate on high-density modes. This mirrors the fidelity--diversity trade-off controlled by classifier-free guidance, but here it comes from a single network with no retraining.}
    \label{fig:sampling_strategy}
\vspace{-0.4cm}
\end{figure}

Sampling from individual heads (left) suggests complementary roles. The early head recovers local texture without global structure (a), the middle head captures object layout without fine detail (b), and the final head preserves coarse semantics alone (c), while their sum (d) satisfies all three simultaneously. Gradient cosine similarity between heads (middle) is high early in sampling and decreases in later steps, suggesting a shared initial direction followed by more specialized contributions.

\textbf{Ablation on Representation Supervision.} As shown in Figure~\ref{fig:fid_comparison}, PRA lowers FID-50K of TEM-B from 42.8 to 19.9 at 50K steps. Additionally aligning negative samples (Pos+Neg RA) still improves over no alignment but reaches only 36.6, erasing most of this gain. As discussed in Section~\ref{sec:training_ebm}, negatives provide a wrong alignment target, and aligning them can even be harmful.

\subsection{Test-Time Scaling Analysis}
\label{sec:test-time}
We investigate the effect of scaling sampling dynamics at inference time by varying both the number of sampling steps and the step size. Figure~\ref{fig:langevin_steps} visualizes the progressive refinement: at $T=40$ (the training horizon), samples contain residual noise, but extended sampling produces increasingly sharp, class-typical images.

Figure~\ref{fig:sampling_strategy} quantifies this trade-off. Increasing the number of steps from 40 to 120 and scaling the step size from 500 to 1500 yields higher Inception Scores (indicating more recognizable samples) at the cost of slightly elevated FID (indicating reduced diversity). Notably, this FID-IS trade-off curve closely resembles that of classifier-free guidance, despite requiring no architectural modifications or unconditional network evaluations, but rather just inference-time scaling.

This controllability lets users tune sampling to the application, with longer, larger-step trajectories for fidelity and shorter, smaller-step ones for diversity.

\begin{figure*}[t]
\centering
\setlength{\tabcolsep}{2pt}
\begin{tabular}{@{}cccccccc@{}}
\includegraphics[width=\dimexpr(\linewidth-28pt)/8\relax]{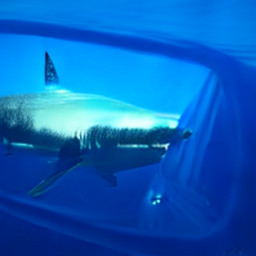} &
\includegraphics[width=\dimexpr(\linewidth-28pt)/8\relax]{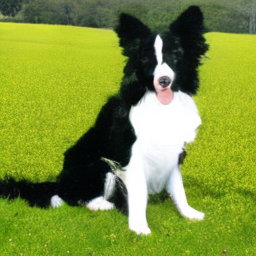}\hspace{3pt}\vrule &
\includegraphics[width=\dimexpr(\linewidth-28pt)/8\relax]{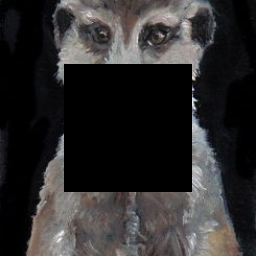} &
\includegraphics[width=\dimexpr(\linewidth-28pt)/8\relax]{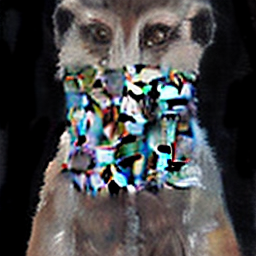} &
\includegraphics[width=\dimexpr(\linewidth-28pt)/8\relax]{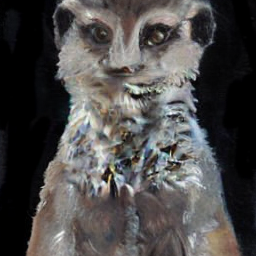} &
\includegraphics[width=\dimexpr(\linewidth-28pt)/8\relax]{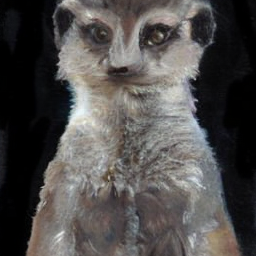} &
\includegraphics[width=\dimexpr(\linewidth-28pt)/8\relax]{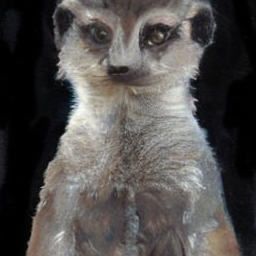} &
\includegraphics[width=\dimexpr(\linewidth-28pt)/8\relax]{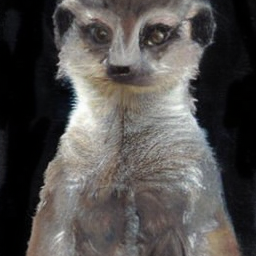} \\[2pt]
\includegraphics[width=\dimexpr(\linewidth-28pt)/8\relax]{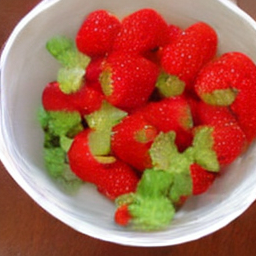} &
\includegraphics[width=\dimexpr(\linewidth-28pt)/8\relax]{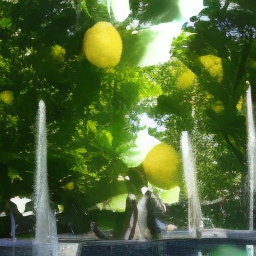}\hspace{3pt}\vrule &
\includegraphics[width=\dimexpr(\linewidth-28pt)/8\relax]{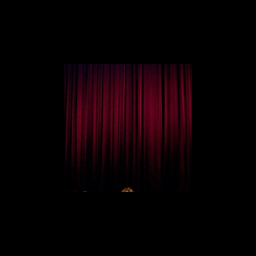} &
\includegraphics[width=\dimexpr(\linewidth-28pt)/8\relax]{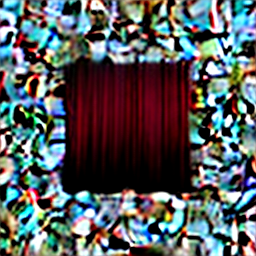} &
\includegraphics[width=\dimexpr(\linewidth-28pt)/8\relax]{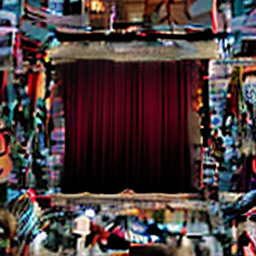} &
\includegraphics[width=\dimexpr(\linewidth-28pt)/8\relax]{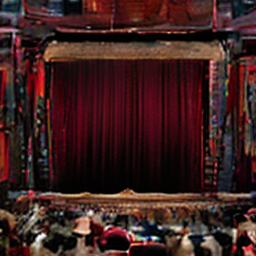} &
\includegraphics[width=\dimexpr(\linewidth-28pt)/8\relax]{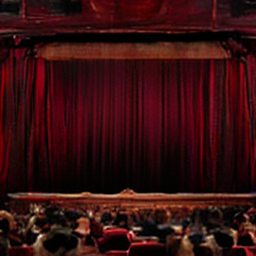} &
\includegraphics[width=\dimexpr(\linewidth-28pt)/8\relax]{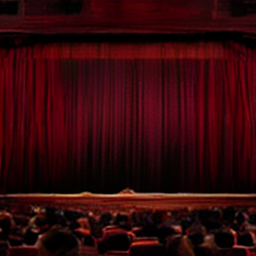} \\[1pt]
\multicolumn{2}{c}{\footnotesize Composed ImageNet} &
\footnotesize Masked & \footnotesize $T=0$ & \footnotesize $T=20$ &
\footnotesize $T=40$ & \footnotesize $T=60$ & \footnotesize $T=80$ \\
\end{tabular}
\vspace{-10pt}
\caption{\textbf{Application of TEM for Image Composition and Inpainting.}
\textit{Left}: composed ImageNet images, car mirror + white shark and border collie + rapeseed (top),
strawberry + petri dish and lemon + fountain (bottom).
\textit{Right}: inpainting of a meerkat (top) and a movie theater (bottom) across sampling steps. Both applications require no retraining or architectural modification.}%
\label{fig:apply}
\vspace{-20pt}
\end{figure*}

\subsection{Applications of TEM}

\setlength{\intextsep}{0pt}
\begin{wrapfigure}[22]{r}{0.45\textwidth}
  \centering
  \setlength{\tabcolsep}{2pt}
  \vspace{-50pt}
  \begin{tabular}{@{}ccc@{}}
    \includegraphics[width=\dimexpr(\linewidth-8pt)/3\relax]{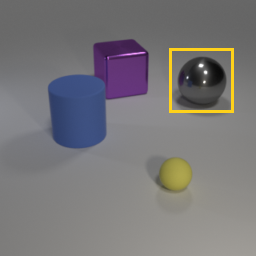} &
    \includegraphics[width=\dimexpr(\linewidth-8pt)/3\relax]{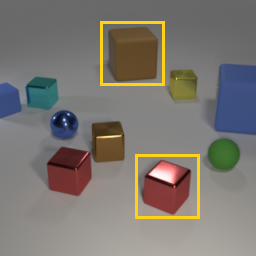} &
    \includegraphics[width=\dimexpr(\linewidth-8pt)/3\relax]{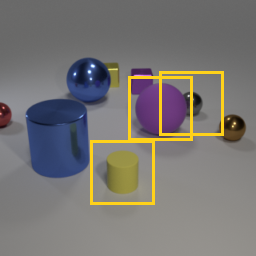} \\
    \includegraphics[width=\dimexpr(\linewidth-8pt)/3\relax]{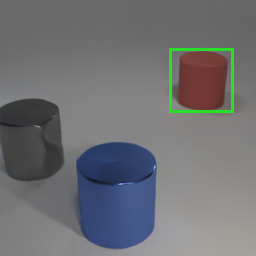} &
    \includegraphics[width=\dimexpr(\linewidth-8pt)/3\relax]{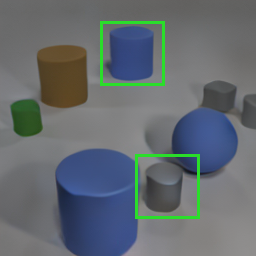} &
    \includegraphics[width=\dimexpr(\linewidth-8pt)/3\relax]{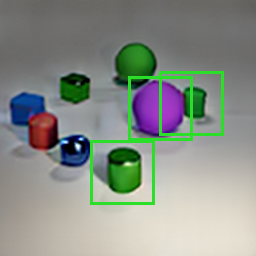} \\
  \end{tabular}\\[4pt]

  \setlength{\tabcolsep}{5pt}
  \renewcommand{\arraystretch}{1.15}
  {\footnotesize
   \begin{tabular*}{\linewidth}{@{\extracolsep{\fill}}lccc@{}}
    \toprule
    \textbf{Models} & \textbf{1 obj.} & \textbf{2 obj.} & \textbf{3 obj.} \\
    \midrule
    Comp. Diff.         & 0.863          & 0.598          & 0.262          \\
    \textbf{TEM (Ours)} & \textbf{0.956} & \textbf{0.766} & \textbf{0.274} \\
    \bottomrule
  \end{tabular*}}
  \caption{\textbf{CLEVR 256$\times$256 compositional generation.}
    \textit{Top:} conditioning inputs, yellow boxes indicate specified object
    positions; second row shows TEM-B samples, green boxes indicate detected
    objects. \textit{Bottom:} compositional generation accuracy ($\uparrow$),
    object detection success rate by a trained classifier over 1000 scenes for
    1-, 2-, and 3-object compositions.}%
  \label{fig:CLEVR}
  \vspace{-25pt}
\end{wrapfigure}

\textbf{Compositional Generation.} Compositional generation arises naturally in the energy-based formulation.
Let $p_1(x) \propto \exp(-E_1(x))$ and $p_2(x) \propto \exp(-E_2(x))$
denote two energy-defined distributions.
Their composition is defined as the product distribution
$p_{\mathrm{comp}}(x) \propto p_1(x)p_2(x)$, which can be equivalently
written as $
p_{\mathrm{comp}}(x) \propto \exp\!\left(-(E_1(x)+E_2(x))\right).
$ Consequently, compositional sampling is obtained by replacing the original energy with the summed energy $E_1(x)+E_2(x)$, enabling joint generation that simultaneously satisfies both
energy functions.

We train TEM-B conditioned on object location and generate compositions of multiple objects at multiple specified positions with CLEVR dataset. Figure~\ref{fig:CLEVR} shows qualitative samples alongside a comparison against Composable Diffusion~\citep{liu2022compositional}, where we report object detection accuracy for 1-, 2-, and 3-object compositions. Following their evaluation protocol, we train the classifier ourselves using their training recipe. TEM outperforms Composable Diffusion at every
composition count. Experimental details are provided in Appendix~\ref{sec:clevr}.

\setlength{\intextsep}{0pt}
\begin{wrapfigure}[7]{r}{0.45\textwidth}
  \centering
  \vspace{-48pt}
  \captionsetup{type=table, aboveskip=9pt, belowskip=0pt}
  \caption{\textbf{Out-of-distribution detection on ImageNet benchmarks
    (AUROC$\uparrow$).} Comparison of DiT-XL/2, SiT-XL/2, SiT-XL/2+REPA and our TEM-XL.}
  \label{tab:ood}
  \setlength{\tabcolsep}{3pt}
  \renewcommand{\arraystretch}{1.15}
  {\footnotesize
  \begin{tabular*}{\linewidth}{@{\extracolsep{\fill}}lcccc@{}}
    \toprule
    \textbf{OOD Dataset} & \textbf{DiT} & \textbf{SiT} &\textbf{SiT-REPA} &\textbf{Ours} \\
    \midrule
    iNaturalist & 0.777 & 0.733&0.780 & \textbf{0.904} \\
    Textures    & 0.439 & 0.426&0.438 & \textbf{0.687} \\
    Places365   & \textbf{0.669} & 0.621&0.481 & 0.636 \\
    SUN         & 0.505 & 0.513&0.500 & \textbf{0.640} \\
    \midrule
    Average     & 0.598 & 0.573&0.549 & \textbf{0.717} \\
    \bottomrule
  \end{tabular*}}
\end{wrapfigure}

\textbf{Out-of-Distribution Detection.}
EBMs naturally provide a density-based score for OOD detection. $E_\theta(x)$ assigns lower energy to in-distribution data and higher energy to OOD inputs~\citep{du2019implicit}. Thus, a single forward pass of $E_\theta(x)$ directly provides an OOD score without any task-specific training. In contrast, diffusion models do not directly provide such a scalar energy and require an additional scoring procedure based on their score function. For the diffusion baselines, we follow~\citet{wyatt2022anoddpm}.
Using the ImageNet-1k validation set as the in-distribution reference, we report AUROC ($\uparrow$) on four standard
benchmarks~\citep{huang2021mos}. Experiment details are in Appendix~\ref{sec:appendix_ood}.

\textbf{Inpainting.}
Inpainting is particularly simple in energy-based models.
Our EBM defines a single energy over image latents,
allowing refinement from any intermediate completion.
We mask the input image before VAE encoding, preventing access to
ground-truth content in the missing region.
We then apply sampling updates to the missing latent region,
restoring the observed-region latents to their initial encoded
values after each step.
Figure~\ref{fig:apply} shows image composition and inpainting results.%

\section{Related Work}

\textbf{Foundations of Energy-Based Models.}
Energy-based models assign an unnormalized scalar energy to each input and define a Gibbs distribution over the data space, an idea with deep roots in statistical physics and undirected graphical models~\citep{hinton2006unsupervised, lecun2006tutorial}.
Classical instantiations such as Restricted Boltzmann Machines relied on tractable architectures to make learning feasible, which in turn limited their representational capacity.
\citet{du2019implicit} showed that short-run Langevin MCMC combined with a replay buffer yields a practical training loop, producing coherent samples on CIFAR-10.
Follow-up work refined the training recipe through improved MCMC initialization, sampler matching terms, and regularization strategies~\citep{du2020improved, nijkamp2019learning, nijkamp2020anatomy}, yet these methods operated at resolutions of 32$\times$32 or less diverse dataset, leaving the scalability of the paradigm unresolved.

\textbf{Latent-Space Energy-Based Models.}
\citet{pang2020learning} proposed training an EBM as a prior over the latent codes of a variational autoencoder, while \citet{hill2022learning} explored energy-based priors for hierarchical latent variable models.
\citet{nie2021controllable} further demonstrated controllable generation by manipulating energies in a learned latent manifold.
VAEBM~\citep{xiao2020vaebm} combines a VAE with an EBM refinement in data space, reparameterizing the MCMC chain through the VAE's latent space to make sampling practical at higher resolutions.
In contrast, we train a single scalar output Transformer energy function in the frozen latent space of a
pretrained autoencoder~\citep{rombach2022high}, keeping the EBM as the sole generative
component.
 
\textbf{Scaling Energy-Based Models.}
Several recent efforts have pushed EBMs toward higher resolution and stronger performance, though each introduces auxiliary mechanisms that depart from the pure energy-based formulation. EGC~\citep{guo2023egc} hybridizes energy-based learning with diffusion modeling and can produce 256$\times$256 ImageNet samples. However, its sampling procedure relies on the full diffusion denoising trajectory.
On the adversarial side, \citet{grathwohl2019your} reinterpret classifier
logits as energies, and \citet{yin2026scalable} scale this idea to ImageNet
256$\times$256. However, their sampler is initialized from an auxiliary image dataset at both training and inference, and is never seeded from noise at this scale.

Trained contrastively, TEM generates high-fidelity 256$\times$256 images, enabling
inference-time scaling and all EBM native applications. Appendix~\ref{app:additionalrelated} provides an extended discussion of related work.

\vspace{-7pt}
\section{Limitations and Conclusion}
\vspace{-3pt}

\textbf{Limitations.} TEM's primary limitation is the computational cost of iterative sampling during both training and inference. One promising direction is to train a feed-forward proposal model to generate candidates in low-energy regions, reducing the sampling needed to refine them.  Extended sampling also introduces a fidelity–diversity trade-off: longer trajectories sharpen images but can reduce diversity and worsen FID. Adaptive noise schedules and proposals that enable movement between distinct modes could help preserve diversity while improving sample quality.

\textbf{Conclusion.} In this work, we introduce the Transformer Energy-Based Model (TEM), an approach that addresses long-standing scalability issues in contrastive training of energy-based models. With a modern transformer backbone, multi-layer energy extraction, and representation-aware training, we demonstrate that a contrastively trained EBM can synthesize high-fidelity ImageNet images at 256$\times$256 and 512$\times$512, competitive with other generative models. We also show that the properties that motivate energy-based modeling in the first place, compositional generation, inpainting, and out-of-distribution detection, remain available at this resolution. We hope this work serves as a starting point for further scaling of contrastively trained EBMs.

\section*{Acknowledgments}
We thank William Moy for helpful discussion and feedback. The authors acknowledge computational support from the Kempner Institute. Jungsoo Lee is supported from Korea Foundation for Advance Studies Fellowship.

\newpage

\bibliography{iclr2027_conference}

@article{hinton2006unsupervised,
  title={Unsupervised discovery of nonlinear structure using contrastive backpropagation},
  author={Hinton, Geoffrey and Osindero, Simon and Welling, Max and Teh, Yee-Whye},
  journal={Cognitive science},
  volume={30},
  number={4},
  pages={725--731},
  year={2006},
  publisher={Wiley Online Library}
}

@article{lecun2006tutorial,
  title={A tutorial on energy-based learning},
  author={LeCun, Yann and Chopra, Sumit and Hadsell, Raia and Ranzato, M and Huang, Fujie and others},
  journal={Predicting structured data},
  volume={1},
  number={0},
  year={2006}
}

@article{du2019implicit,
  title={Implicit generation and modeling with energy based models},
  author={Du, Yilun and Mordatch, Igor},
  journal={Advances in neural information processing systems},
  volume={32},
  year={2019}
}

@article{du2020improved,
  title={Improved contrastive divergence training of energy based models},
  author={Du, Yilun and Li, Shuang and Tenenbaum, Joshua and Mordatch, Igor},
  journal={arXiv preprint arXiv:2012.01316},
  year={2020}
}

@article{beirlant1997nonparametric,
  title={Nonparametric entropy estimation: An overview},
  author={Beirlant, Jan and Dudewicz, Edward J and Gy{\"o}rfi, L{\'a}szl{\'o} and Van der Meulen, Edward C and others},
  journal={International Journal of Mathematical and Statistical Sciences},
  volume={6},
  number={1},
  pages={17--39},
  year={1997},
  publisher={THESAURUS PUBLISHING}
}

@article{wang2025equilibrium,
  title={Equilibrium matching: Generative modeling with implicit energy-based models},
  author={Wang, Runqian and Du, Yilun},
  journal={arXiv preprint arXiv:2510.02300},
  year={2025}
}

@inproceedings{guo2023egc,
  title={Egc: Image generation and classification via a diffusion energy-based model},
  author={Guo, Qiushan and Ma, Chuofan and Jiang, Yi and Yuan, Zehuan and Yu, Yizhou and Luo, Ping},
  booktitle={Proceedings of the IEEE/CVF International Conference on Computer Vision},
  pages={22952--22962},
  year={2023}
}

@article{grathwohl2019your,
  title={Your classifier is secretly an energy based model and you should treat it like one},
  author={Grathwohl, Will and Wang, Kuan-Chieh and Jacobsen, J{\"o}rn-Henrik and Duvenaud, David and Norouzi, Mohammad and Swersky, Kevin},
  journal={arXiv preprint arXiv:1912.03263},
  year={2019}
}

@article{hill2022learning,
  title={Learning probabilistic models from generator latent spaces with hat ebm},
  author={Hill, Mitch and Nijkamp, Erik and Mitchell, Jonathan and Pang, Bo and Zhu, Song-Chun},
  journal={Advances in Neural Information Processing Systems},
  volume={35},
  pages={928--940},
  year={2022}
}

@article{nijkamp2019learning,
  title={Learning non-convergent non-persistent short-run mcmc toward energy-based model},
  author={Nijkamp, Erik and Hill, Mitch and Zhu, Song-Chun and Wu, Ying Nian},
  journal={Advances in Neural Information Processing Systems},
  volume={32},
  year={2019}
}

@article{pang2020learning,
  title={Learning latent space energy-based prior model},
  author={Pang, Bo and Han, Tian and Nijkamp, Erik and Zhu, Song-Chun and Wu, Ying Nian},
  journal={Advances in Neural Information Processing Systems},
  volume={33},
  pages={21994--22008},
  year={2020}
}

@article{nie2021controllable,
  title={Controllable and compositional generation with latent-space energy-based models},
  author={Nie, Weili and Vahdat, Arash and Anandkumar, Anima},
  journal={Advances in Neural Information Processing Systems},
  volume={34},
  pages={13497--13510},
  year={2021}
}

@article{xiao2020vaebm,
  title={Vaebm: A symbiosis between variational autoencoders and energy-based models},
  author={Xiao, Zhisheng and Kreis, Karsten and Kautz, Jan and Vahdat, Arash},
  journal={arXiv preprint arXiv:2010.00654},
  year={2020}
}

@inproceedings{nijkamp2020anatomy,
  title={On the anatomy of mcmc-based maximum likelihood learning of energy-based models},
  author={Nijkamp, Erik and Hill, Mitch and Han, Tian and Zhu, Song-Chun and Wu, Ying Nian},
  booktitle={Proceedings of the AAAI Conference on Artificial Intelligence},
  volume={34},
  number={04},
  pages={5272--5280},
  year={2020}
}

@article{dosovitskiy2020image,
  title={An image is worth 16x16 words: Transformers for image recognition at scale},
  author={Dosovitskiy, Alexey and Beyer, Lucas and Kolesnikov, Alexander and Weissenborn, Dirk and Zhai, Xiaohua and Unterthiner, Thomas and Dehghani, Mostafa and Minderer, Matthias and Heigold, Georg and Gelly, Sylvain and others},
  journal={arXiv preprint arXiv:2010.11929},
  year={2020}
}

@inproceedings{peebles2023scalable,
  title={Scalable diffusion models with transformers},
  author={Peebles, William and Xie, Saining},
  booktitle={Proceedings of the IEEE/CVF international conference on computer vision},
  pages={4195--4205},
  year={2023}
}

@article{su2024roformer,
  title={Roformer: Enhanced transformer with rotary position embedding},
  author={Su, Jianlin and Ahmed, Murtadha and Lu, Yu and Pan, Shengfeng and Bo, Wen and Liu, Yunfeng},
  journal={Neurocomputing},
  volume={568},
  pages={127063},
  year={2024},
  publisher={Elsevier}
}

@article{shazeer2020glu,
  title={Glu variants improve transformer},
  author={Shazeer, Noam},
  journal={arXiv preprint arXiv:2002.05202},
  year={2020}
}

@inproceedings{henry2020query,
  title={Query-key normalization for transformers},
  author={Henry, Alex and Dachapally, Prudhvi Raj and Pawar, Shubham Shantaram and Chen, Yuxuan},
  booktitle={Findings of the Association for Computational Linguistics: EMNLP 2020},
  pages={4246--4253},
  year={2020}
}

@inproceedings{rombach2022high,
  title={High-resolution image synthesis with latent diffusion models},
  author={Rombach, Robin and Blattmann, Andreas and Lorenz, Dominik and Esser, Patrick and Ommer, Bj{\"o}rn},
  booktitle={Proceedings of the IEEE/CVF conference on computer vision and pattern recognition},
  pages={10684--10695},
  year={2022}
}

@article{oquab2024dinov2,
  title={DINOv2: Learning robust visual features without supervision},
  author={Oquab, Maxime and Darcet, Timoth{\'e}e and Moutakanni, Th{\'e}o and Vo, Huy and Szafraniec, Marc and Khalidov, Vasil and Fernandez, Pierre and Haziza, Daniel and Massa, Francisco and El-Nouby, Alaaeldin and others},
  journal={Transactions on Machine Learning Research},
  year={2024}
}

@article{wang2025diffuse,
  title={Diffuse and disperse: Image generation with representation regularization},
  author={Wang, Runqian and He, Kaiming},
  journal={arXiv preprint arXiv:2506.09027},
  year={2025}
}

@article{yu2024representation,
  title={Representation alignment for generation: Training diffusion transformers is easier than you think},
  author={Yu, Sihyun and Kwak, Sangkyung and Jang, Huiwon and Jeong, Jongheon and Huang, Jonathan and Shin, Jinwoo and Xie, Saining},
  journal={arXiv preprint arXiv:2410.06940},
  year={2024}
}

@article{singh2025matters,
  title={What matters for Representation Alignment: Global Information or Spatial Structure?},
  author={Singh, Jaskirat and Leng, Xingjian and Wu, Zongze and Zheng, Liang and Zhang, Richard and Shechtman, Eli and Xie, Saining},
  journal={arXiv preprint arXiv:2512.10794},
  year={2025}
}

@article{dhariwal2021diffusion,
  title={Diffusion models beat gans on image synthesis},
  author={Dhariwal, Prafulla and Nichol, Alexander},
  journal={Advances in neural information processing systems},
  volume={34},
  pages={8780--8794},
  year={2021}
}

@article{raghu2021vision,
  title={Do vision transformers see like convolutional neural networks?},
  author={Raghu, Maithra and Unterthiner, Thomas and Kornblith, Simon and Zhang, Chiyuan and Dosovitskiy, Alexey},
  journal={Advances in neural information processing systems},
  volume={34},
  pages={12116--12128},
  year={2021}
}

@inproceedings{yin2026scalable,
  title={Scalable Energy-Based Models via Adversarial Training: Unifying Discrimination and Generation},
  author={Yin, Xuwang and Zhang, Claire and Steele, Julie and Shavit, Nir and Wang, Tony},
  booktitle={International Conference on Learning Representations},
  volume={2026},
  pages={68385--68424},
  year={2026}
}

@inproceedings{ma2024sit,
  title={SiT: Exploring flow and diffusion-based generative models with scalable interpolant transformers},
  author={Ma, Nanye and Goldstein, Mark and Albergo, Michael S and Boffi, Nicholas M and Vanden-Eijnden, Eric and Xie, Saining},
  booktitle={European Conference on Computer Vision},
  pages={23--40},
  year={2024},
  organization={Springer}
}

@inproceedings{
brock2018large,
title={Large Scale {GAN} Training for High Fidelity Natural Image Synthesis},
author={Andrew Brock and Jeff Donahue and Karen Simonyan},
booktitle={International Conference on Learning Representations},
year={2019},
url={https://openreview.net/forum?id=B1xsqj09Fm},
}

@inproceedings{kang2023scaling,
  title={Scaling up gans for text-to-image synthesis},
  author={Kang, Minguk and Zhu, Jun-Yan and Zhang, Richard and Park, Jaesik and Shechtman, Eli and Paris, Sylvain and Park, Taesung},
  booktitle={Proceedings of the IEEE/CVF conference on computer vision and pattern recognition},
  pages={10124--10134},
  year={2023}
}

@inproceedings{deng2009imagenet,
  title={Imagenet: A large-scale hierarchical image database},
  author={Deng, Jia and Dong, Wei and Socher, Richard and Li, Li-Jia and Li, Kai and Fei-Fei, Li},
  booktitle={2009 IEEE conference on computer vision and pattern recognition},
  pages={248--255},
  year={2009},
  organization={Ieee}
}

@inproceedings{johnson2017clevr,
  title={Clevr: A diagnostic dataset for compositional language and elementary visual reasoning},
  author={Johnson, Justin and Hariharan, Bharath and Van Der Maaten, Laurens and Fei-Fei, Li and Lawrence Zitnick, C and Girshick, Ross},
  booktitle={Proceedings of the IEEE conference on computer vision and pattern recognition},
  pages={2901--2910},
  year={2017}
}

@inproceedings{du2023reduce,
  title={Reduce, reuse, recycle: Compositional generation with energy-based diffusion models and mcmc},
  author={Du, Yilun and Durkan, Conor and Strudel, Robin and Tenenbaum, Joshua B and Dieleman, Sander and Fergus, Rob and Sohl-Dickstein, Jascha and Doucet, Arnaud and Grathwohl, Will Sussman},
  booktitle={International conference on machine learning},
  pages={8489--8510},
  year={2023},
  organization={PMLR}
}

@inproceedings{liu2022compositional,
  title={Compositional visual generation with composable diffusion models},
  author={Liu, Nan and Li, Shuang and Du, Yilun and Torralba, Antonio and Tenenbaum, Joshua B},
  booktitle={European conference on computer vision},
  pages={423--439},
  year={2022},
  organization={Springer}
}

@article{tian2024visual,
  title={Visual autoregressive modeling: Scalable image generation via next-scale prediction},
  author={Tian, Keyu and Jiang, Yi and Yuan, Zehuan and Peng, Bingyue and Wang, Liwei},
  journal={Advances in neural information processing systems},
  volume={37},
  pages={84839--84865},
  year={2024}
}

@inproceedings{chang2022maskgit,
  title={Maskgit: Masked generative image transformer},
  author={Chang, Huiwen and Zhang, Han and Jiang, Lu and Liu, Ce and Freeman, William T},
  booktitle={Proceedings of the IEEE/CVF conference on computer vision and pattern recognition},
  pages={11315--11325},
  year={2022}
}

@inproceedings{wyatt2022anoddpm,
  title={Anoddpm: Anomaly detection with denoising diffusion probabilistic models using simplex noise},
  author={Wyatt, Julian and Leach, Adam and Schmon, Sebastian M and Willcocks, Chris G},
  booktitle={2022 IEEE/CVF Conference on Computer Vision and Pattern Recognition Workshops (CVPRW)},
  pages={649--655},
  year={2022},
  organization={IEEE}
}

@inproceedings{huang2021mos,
  title={Mos: Towards scaling out-of-distribution detection for large semantic space},
  author={Huang, Rui and Li, Yixuan},
  booktitle={Proceedings of the IEEE/CVF conference on computer vision and pattern recognition},
  pages={8710--8719},
  year={2021}
}

@inproceedings{xiao2010sun,
  title={Sun database: Large-scale scene recognition from abbey to zoo},
  author={Xiao, Jianxiong and Hays, James and Ehinger, Krista A and Oliva, Aude and Torralba, Antonio},
  booktitle={2010 IEEE computer society conference on computer vision and pattern recognition},
  pages={3485--3492},
  year={2010},
  organization={IEEE}
}

@inproceedings{van2018inaturalist,
  title={The inaturalist species classification and detection dataset},
  author={Van Horn, Grant and Mac Aodha, Oisin and Song, Yang and Cui, Yin and Sun, Chen and Shepard, Alex and Adam, Hartwig and Perona, Pietro and Belongie, Serge},
  booktitle={Proceedings of the IEEE conference on computer vision and pattern recognition},
  pages={8769--8778},
  year={2018}
}

@article{zhou2017places,
  title={Places: A 10 million image database for scene recognition},
  author={Zhou, Bolei and Lapedriza, Agata and Khosla, Aditya and Oliva, Aude and Torralba, Antonio},
  journal={IEEE transactions on pattern analysis and machine intelligence},
  volume={40},
  number={6},
  pages={1452--1464},
  year={2017},
  publisher={IEEE}
}

@inproceedings{cimpoi2014describing,
  title={Describing textures in the wild},
  author={Cimpoi, Mircea and Maji, Subhransu and Kokkinos, Iasonas and Mohamed, Sammy and Vedaldi, Andrea},
  booktitle={Proceedings of the IEEE conference on computer vision and pattern recognition},
  pages={3606--3613},
  year={2014}
}

@article{lipman2022flow,
  title={Flow matching for generative modeling},
  author={Lipman, Yaron and Chen, Ricky TQ and Ben-Hamu, Heli and Nickel, Maximilian and Le, Matt},
  journal={arXiv preprint arXiv:2210.02747},
  year={2022}
}

@inproceedings{sohl2015deep,
  title={Deep unsupervised learning using nonequilibrium thermodynamics},
  author={Sohl-Dickstein, Jascha and Weiss, Eric and Maheswaranathan, Niru and Ganguli, Surya},
  booktitle={International conference on machine learning},
  pages={2256--2265},
  year={2015},
  organization={pmlr}
}

@article{ho2020denoising,
  title={Denoising diffusion probabilistic models},
  author={Ho, Jonathan and Jain, Ajay and Abbeel, Pieter},
  journal={Advances in neural information processing systems},
  volume={33},
  pages={6840--6851},
  year={2020}
}

@article{balcerak2025energy,
  title={Energy matching: Unifying flow matching and energy-based models for generative modeling},
  author={Balcerak, Michal and Amiranashvili, Tamaz and Terpin, Antonio and Shit, Suprosanna and Bogensperger, Lea and Kaltenbach, Sebastian and Koumoutsakos, Petros and Menze, Bjoern},
  journal={Advances in Neural Information Processing Systems},
  volume={38},
  pages={8583--8609},
  year={2025}
}

@article{gladstone2025energy,
  title={Energy-based transformers are scalable learners and thinkers},
  author={Gladstone, Alexi and Nanduru, Ganesh and Islam, Md Mofijul and Han, Peixuan and Ha, Hyeonjeong and Chadha, Aman and Du, Yilun and Ji, Heng and Li, Jundong and Iqbal, Tariq},
  journal={arXiv preprint arXiv:2507.02092},
  year={2025}
}

@article{kaplan2020scaling,
  title={Scaling laws for neural language models},
  author={Kaplan, Jared and McCandlish, Sam and Henighan, Tom and Brown, Tom B and Chess, Benjamin and Child, Rewon and Gray, Scott and Radford, Alec and Wu, Jeffrey and Amodei, Dario},
  journal={arXiv preprint arXiv:2001.08361},
  year={2020}
}

@article{mildenhall2021nerf,
  title={Nerf: Representing scenes as neural radiance fields for view synthesis},
  author={Mildenhall, Ben and Srinivasan, Pratul P and Tancik, Matthew and Barron, Jonathan T and Ramamoorthi, Ravi and Ng, Ren},
  journal={Communications of the ACM},
  volume={65},
  number={1},
  pages={99--106},
  year={2021},
  publisher={ACM New York, NY, USA}
}
\bibliographystyle{iclr2027_conference}
\newpage
\appendix

\section{The Sampler-Matching Loss}
\label{sec:derivationkl}

Maximum likelihood training of an EBM minimizes
$\mathcal{L}_{\mathrm{NLL}} = \mathbb{E}_{p_{\mathcal{D}}}[E_\theta(x)] + \log Z(\theta)$,
whose gradient is given in Eq.~\ref{eq:mle-grad}: a positive phase over data and
a negative phase over samples from $p_\theta$. The negative phase is the obstacle,
since exact samples from $p_\theta$ require MCMC run to convergence.

In practice we replace $p_\theta$ by the distribution $q_\theta$ of the final state
of a short Langevin chain initialized at $\mathcal{N}(0,I)$ or at a past sample. The resulting gradient,
\begin{equation}
\mathcal{L}_{\mathrm{contrast-only}} = \mathbb{E}_{p_{\mathcal{D}}(x)}[E_\theta(x)]
- \mathbb{E}_{\mathrm{sg}(q_\theta(x'))}[E_\theta(x')],
\end{equation}
is therefore biased by the gap between $q_\theta$ and $p_\theta$, and the bias is
largest exactly when the sampler fails to reach the model's low-energy regions.

Following~\citet{du2020improved}, we introduce a loss
motivated by this gap that encourages low-energy,
diverse sampler outputs:
\begin{equation}
\mathcal{L}_{\mathrm{KL}} = \mathrm{KL}(q_\theta \,\|\, p_{\mathrm{sg}(\theta)})
+ \text{const}
= \mathbb{E}_{q_\theta(x)}[E_{\mathrm{sg}(\theta)}(x)]
+ \mathbb{E}_{q_\theta(x)}[\log q_\theta(x)],
\label{eq:app-lkl}
\end{equation}
where the stop-gradient $\mathrm{sg}(\cdot)$ freezes the energy parameters so that
gradients flow only through the sampler, and the constant
$-\log Z(\mathrm{sg}(\theta))$ does not depend on the differentiated parameters.
Minimizing $\mathcal{L}_{\mathrm{KL}}$ makes the sampler's outputs both low-energy
and diverse, which is precisely what an accurate negative phase requires. It also
counteracts the failure mode in which the energy function shapes a landscape that
its own sampler cannot descend.

\subsection{Estimating the sampler-energy term}
\label{app:kl-lopt}
 
Eq.~\ref{eq:app-lkl} splits into two pieces,
$\mathcal{L}_{\mathrm{KL}} = \mathcal{L}_{\mathrm{opt}} + \mathcal{L}_{\mathrm{ent}}$, with
$\mathcal{L}_{\mathrm{opt}} = \mathbb{E}_{q_\theta(x)}[E_{\mathrm{sg}(\theta)}(x)]$ and
$\mathcal{L}_{\mathrm{ent}} = \mathbb{E}_{q_\theta(x)}[\log q_\theta(x)] = -\mathbb{H}(q_\theta)$.
 
The first is directly tractable: each sampling step is a
differentiable function of $\theta$, so
\begin{equation}
\frac{\partial \mathcal{L}_{\mathrm{opt}}}{\partial \theta}
=
\mathbb{E}_{q_\theta\left(x^{0},\dots,x^{T}\right)}
\left[
\frac{\partial\, E_{\mathrm{sg}(\theta)}\!\left(
x^{T-1} - \eta \nabla_{x} E_\theta\!\left(x^{T-1}\right) + \omega^{T}
\right)}{\partial \theta}
\right],
\label{eq:app-lopt}
\end{equation}
which is a second-order gradient term. Backpropagating through all $T$ steps is
memory-prohibitive at our sequence lengths, so we differentiate only through the final sampling
step, following~\citep{du2020improved}.

\subsection{Estimating the entropy term}
\label{app:kl-lent}
 
The second piece is the obstacle. Evaluating $\log q_\theta(x)$ exactly is
intractable: $q_\theta$ is the pushforward of the buffer distribution through
$T$ sampling steps, and its density at a point requires
marginalizing over all trajectories that could have produced it.

We instead use the non-parametric nearest-neighbor entropy
estimator~\citep{beirlant1997nonparametric}, which estimates $\mathbb{H}(p)$ from
$n$ samples by their average log-distance to the nearest other sample. Up to
factors independent of $\theta$, minimizing $\mathcal{L}_{\mathrm{ent}}$
therefore amounts to
\begin{equation}
\mathcal{L}_{\mathrm{ent}}
\;=\;
\mathbb{E}_{q_\theta(x)}\!\left[-\log \mathrm{NN}\!\left(x, \mathcal{B}\right)\right],
\qquad
\mathrm{NN}\!\left(x, \mathcal{B}\right) = \min_{x' \in \mathcal{B}} \left\| x - x' \right\|_2 ,
\label{eq:app-lent}
\end{equation}
where $\mathcal{B}$ is a set of past MCMC samples treated as constants
(stop-gradient) and distances are $L_2$ in latent space. Repelling samples from
one another increases their diversity.  This class of estimators scale poorly with
dimension, and the buffer samples are neither independent nor drawn from the
current $q_\theta$. As in~\citet{du2020improved}, we use it only as a diversity
regularizer rather than as an accurate estimate of $\mathbb{H}(q_\theta)$.

\section{Training \& Inference Details }
\label{sec:trainingdetail}
\subsection{Training Details.} 

The energy head consists of a two-layer MLP with SiLU activations. Details of the transformer architecture are provided in Table~\ref{tab:dit_scaling}. All experiments operate in the latent space of a pre-trained Stable Diffusion VAE \citep{rombach2022high}, following \citet{peebles2023scalable} and are trained with bfloat16 mixed precision. The latent space is clipped by norm 100 to stabilize sampling for 256$\times$256 and by norm 200 for 512$\times$512, which is chosen as the threshold containing over $99.99\%$ of ImageNet training latents, which stabilizes sampling. A smaller threshold clips real data points, while a larger one expands the region the sampler has to search, which slightly degrades performance.
For sampling, we use a step size of 500 and noise scale of 0.05. With a small step size, the network would have to produce very large input gradients to move samples meaningfully within a limited number of steps, and we found that this destabilizes training. With an excessively large step size, however, each update is coarse, and the model struggles to learn a fine-grained energy landscape. Following prior work~\citep{du2019implicit,du2020improved,gladstone2025energy}, who likewise decouple the step size from the noise scale and use a step size much larger than the standard Langevin prescription, we select the step size empirically to balance these two effects. During training, 40-step sampling is employed. 

The replay buffer's replay probability is set to 0.99 and capacity is 16000. Class labels are drawn independently for every chain, including chains initialized from the buffer. Buffer states therefore serve as class independent initializations near the data manifold, which the conditional energy refines toward the requested class. Once full, each newly added sample replaces a uniformly randomly selected buffer entry. For the sampler matching term, the distance with the nearest neighbor term is calculated based on the distance from 100 randomly sampled seeds to the replay buffer following~\citep{du2020improved}. We use DINOv2-large~\citep{oquab2024dinov2} as the VFM and implement the projection network $P_\phi$ as a lightweight CNN with spatial normalization following~\citet{singh2025matters} attached after the block nearest to one-third of the total network depth. We use $\lambda=1$ for energy L2 regularization, $\lambda_1=1$ and $\lambda_2=0.6$ for overall objective's coefficient at Equation \ref{eq:overall} following \citet{du2020improved} and \citet{singh2025matters}. We optimize the model using Adam with a learning rate of
$2 \times 10^{-4}$, momentum parameters
$(\beta_1, \beta_2) = (0, 0.999)$, $\epsilon = 10^{-8}$, and no weight decay. We keep an exponential moving average (EMA) of the weights with decay $0.9999$,
updated after every optimizer step. All reported samples and metrics use the EMA weights. During training, we replace the class label with a null token
with probability $0.15$, enabling unconditional energy
evaluation for OOD detection.
\begin{table}[h]
\centering
\small
\vspace{10pt}
\caption{\textbf{TEM architecture configurations.}
Model variants from TEM-S to TEM-XL. All variants use a patch size of $2\times2$ and operate in the latent space of a pretrained Stable Diffusion VAE. Scaling from S to XL increases both depth and hidden dimension, yielding monotonic FID improvements (see Figure~\ref{fig:diff_vs_ebm}).}
\label{tab:dit_scaling}
\setlength{\tabcolsep}{6pt}
\renewcommand{\arraystretch}{1.0}

\begin{tabular}{lcccc}
\toprule
\textbf{Model} & \textbf{TEM-S} & \textbf{TEM-B} & \textbf{TEM-L} & \textbf{TEM-XL} \\
\midrule

Params & 33M & 122M & 415M & 609M \\
Depth (layers) & 12 & 12 & 24 & 28 \\
Hidden size & 384 & 768 & 1024 & 1152 \\
Attention heads & 6 & 12 & 16 & 16 \\
\bottomrule
\end{tabular}
\vspace{2pt}

\end{table}

\subsection{Inference details.} Evaluation uses persistent chains: the first 5{,}000 chains are initialized from $\mathcal{N}(\mathbf{0}, \mathbf{I})$, and subsequent chains continue from buffered 40-step states (Algorithm~\ref{alg:langevin}), so no external data enters sampling. Once full, each newly added 40 step state replaces a uniformly randomly selected buffer entry. Class labels are sampled independently for each chain and are randomly mixed throughout evaluation. We apply the same latent-norm clipping threshold used during training. Each output requires $T$ additional sampling steps, including
for noise-initialized chains, giving an amortized cost of $T$ steps per sample under this persistent evaluation protocol.

To discuss more about details about fidelity-diversity trade-off at Section~\ref{sec:test-time}, Table~\ref{tab:steps_pr} directly quantifies this trade-off:
increasing sampling steps from 40 to 80 raises precision
from 0.65 to 0.81 while reducing recall from 0.65 to 0.51. This trend is similar to the fidelity-diversity
trade-off observed with increasing classifier-free guidance
in diffusion models.

\begin{table}[t]
  \centering
  \caption{Fidelity-diversity trade-off. Top: TEM-XL with varying numbers of sampling steps, precision/recall with $k=3$. Bottom: DiT-XL/2  with varying classifier-free guidance scale $g$ results from \citet{peebles2023scalable} }
  \label{tab:steps_pr}
  \begin{tabular}{lccccc}
    \toprule
    Model & \# Steps & FID $\downarrow$ & IS $\uparrow$ & Precision $\uparrow$ & Recall $\uparrow$ \\
    \midrule
    TEM-XL & 40 & 7.63 & 106.0 & 0.65 & 0.65 \\
    TEM-XL & 60 & 3.82 & 176.2 & 0.76 & 0.57 \\
    TEM-XL & 80 & 4.72 & 221.1 & 0.81 & 0.51 \\
    \midrule[\heavyrulewidth]
    DiT-XL ($g$=1.0)  & 250 & 9.62 & 121.5 & 0.67 & 0.67 \\
    DiT-XL ($g$=1.25) & 250 & 3.22 & 201.8 & 0.76 & 0.62 \\
    DiT-XL ($g$=1.5)  & 250 & 2.27 & 278.2 & 0.83 & 0.57 \\
    \bottomrule
  \end{tabular}
\end{table}

\section{Computational Cost}
\label{app:cost}
Training the TEM-XL model for 50K steps on ImageNet 256$\times$256 dataset takes approximately 6 days using 4 H200 GPUs. For sampling, it takes about an hour to sample 50K images with 1 H200 GPU for 60 sampling steps. Higher resolution training on 512$\times$512 takes roughly 3--4 times longer than 256$\times$256 on both training and inference.

\subsection{Training Cost}
\paragraph{Training cost.} We follow the FLOP accounting convention
of~\citet{kaplan2020scaling}, letting $F$ denote the cost of a single forward
pass. A standard training step for a diffusion transformer costs $3F$ per image:
one forward pass and a backward pass of roughly $2F$, since gradients are taken
with respect to both the activations and the parameters. A TEM step has four
components:

\begin{itemize}
\item \textbf{Negative sampling.} Each of the $T_{\text{train}}{=}40$ sampling steps
requires $\nabla_{\mathbf{z}} E_\theta(\mathbf{z})$, a gradient with respect to
the input only, which costs approximately $F$ in addition to the forward pass.
The chain therefore costs about $2FT_{\text{train}} = 80F$.
\item \textbf{Sampler matching loss.} This term calculates the second-order gradient for last sampling step which takes about $8F$ based on measurement.
\item \textbf{Contrastive loss and parameter update.} The energies of the positive and
negative batches are each evaluated and backpropagated through, about $3F$ each.
\item \textbf{Representation alignment.} One forward pass of the frozen VFM,
which is not backpropagated through. For DINOv2-L at our token count this is
approximately $0.9F$.

\end{itemize}

Despite this per-step overhead, TEM reaches a lower guidance-free FID with less
total training compute, because it converges in far fewer steps. Table~\ref{tab:training-cost} compares the total training cost of TEM with that of representative models.

\begin{table}[h]
\centering
\vspace{5pt}
\caption{\textbf{Training cost on ImageNet 256$\times$256.} All values use the same
accounting convention. VFM forwards are included for TEM and SiT+REPA.}
\label{tab:training-cost}
\begin{tabular}{lccc}
\toprule
Model & Training steps & Training FLOPs & FID $\downarrow$ \\
\midrule
DiT-XL/2 & 7M & 638 E & 9.62 \\
SiT-XL/2 & 7M & 638 E & 8.30 \\
SiT-XL/2+REPA & 4M & 393 E & 5.90 \\
TEM-XL & 50K & 146 E & 3.82 \\
\bottomrule
\end{tabular}
\end{table}

\subsection{Inference Cost}
\label{sec:flops}

Because each sampling requires the input gradient $\nabla_{\mathbf{z}}
E_\theta(\mathbf{z})$ at every step, the number of steps is not
directly comparable to that of diffusion models, we report FLOPs instead.
Following the accounting convention of~\citet{kaplan2020scaling}, let $F$ denote
the cost of one forward pass. Sampling requires only the gradient with
respect to the input, which costs approximately $F$, as opposed to roughly $2F$
when parameter gradients are also required. Each sampling step therefore costs
about $2F$, and sampling with $K$ steps costs $2FK$. Table~\ref{tab:sampling-flops} compares the sampling cost of TEM with
that of other models.

\begin{table}[h]
\centering
\vspace{5pt}
\caption{\textbf{Sampling cost on ImageNet 256$\times$256.}  All results are
without classifier-free guidance. }
\label{tab:sampling-flops}
\begin{tabular}{lccc}
\toprule
Model & Steps / Sample & FLOPs / Sample & FID $\downarrow$ \\
\midrule
DiT-XL/2 & 250 & 29.66 T & 9.62 \\
SiT-XL/2 & 250 & 29.66 T & 8.30 \\
TEM-L    & 60  & 9.87 T & 4.74 \\
TEM-XL   & 60  & 14.48 T & 3.82 \\
\bottomrule
\end{tabular}
\end{table}

Although each sampling step costs roughly twice the cost of a single forward pass, TEM uses
far fewer steps while reaching a lower guidance-free FID. Classifier-free guidance
would improve the baselines' quality but doubles their sampling cost. Although accelerated solvers can further reduce the cost of diffusion sampling, the overhead of computing input gradients is modest in practice, and TEM attains a lower guidance-free FID at a comparable sampling cost.
\section{Training Algorithm}
\label{training algorithm}
The full training procedure for TEM is given in Algorithm~\ref{alg:training}.
\begin{algorithm}[t]
\caption{Transformer Energy-Based Model (TEM) Training}
\label{alg:training}
\begin{algorithmic}[1]
\State {\bfseries Input:} Data distribution $p_{D}(\mathbf{x})$, encoder $\mathcal{E}$, frozen VFM $h_{\mathrm{VFM}}$, sampling step size $\eta$, noise scale $\sigma$, number of steps $T$, loss coefficients $\lambda, \lambda_1, \lambda_2$, replay probability $p_{\text{replay}}$
\State {\bfseries Initialize:} Replay buffer $\mathcal{B} \leftarrow \emptyset$
\While{not converged}
    \State \textcolor{gray}{\# 1. Sample positive data and initialize negative chains}
    \State Sample real images $\mathbf{x}^{+} \sim p_{D}(\mathbf{x})$; $\quad \mathbf{z}^{+} \leftarrow \mathcal{E}(\mathbf{x}^{+})$
    \State Sample $\tilde{\mathbf{z}}^{0} \sim \mathcal{B}$ with probability $p_{\text{replay}}$, else $\tilde{\mathbf{z}}^{0} \sim \mathcal{N}(0, I)$
    \State \textcolor{gray}{\# 2. Sampling}
    \For{$t = 1$ {\bfseries to} $T$}
        \State $\tilde{\mathbf{z}}^{t} \leftarrow \tilde{\mathbf{z}}^{t-1} - \eta\nabla_{\mathbf{z}}E_{\theta}(\tilde{\mathbf{z}}^{t-1}) + \omega, \quad \omega \sim \mathcal{N}(0, \sigma^2)$
    \EndFor
    \State $\hat{\mathbf{z}}^{-} \leftarrow \tilde{\mathbf{z}}^{T}$; $\quad \mathbf{z}^{-} \leftarrow \mathrm{sg}(\tilde{\mathbf{z}}^{T})$ \Comment{with and without gradient}
    \State \textcolor{gray}{\# 3. Compute Training Objectives}
    \State $\mathcal{L} \leftarrow \frac{1}{N}\sum \Big( E_{\theta}(\mathbf{z}^{+}) - E_{\theta}(\mathbf{z}^{-}) + \lambda\big(E_{\theta}(\mathbf{z}^{+})^{2} + E_{\theta}(\mathbf{z}^{-})^{2}\big)$
    \State $\qquad\qquad\quad + \lambda_1\big(E_{\mathrm{sg}(\theta)}(\hat{\mathbf{z}}^{-}) - \log \mathrm{NN}(\hat{\mathbf{z}}^{-}, \mathcal{B})\big) + \lambda_2 \mathcal{L}_{\mathrm{PRA}}(\mathbf{x}^{+}, \mathbf{z}^{+}) \Big)$
    \State \textcolor{gray}{\# 4. Backpropagation and Parameter Update}
    \State $\Delta\theta \leftarrow \nabla_{\theta}\mathcal{L}$
    \State \textcolor{gray}{\# 5. Update persistent buffer}
    \State $\mathcal{B} \leftarrow \mathcal{B} \cup \{\mathbf{z}^{-}\}$
\EndWhile
\end{algorithmic}
\end{algorithm}

\section{Compositional Generation in 256x256 CLEVR Dataset}
\label{sec:clevr}
To quantitatively evaluate TEM’s compositional performance, we train TEM on a 256×256 location-conditioned CLEVR~\citep{johnson2017clevr} dataset consisting of 70,000 images. Prior work on compositional
generation evaluates at 128$\times$128 with a classifier trained at that
resolution. We extend this setting to 256×256 to test whether the compositional advantage of energy-based models persists at higher resolution, which is a harder generative setting. The
following describes the training setup and the evaluation protocol used to
compare against Composable Diffusion~\citep{liu2022compositional}.

\textbf{Training a location-conditioned TEM on CLEVR 256$\times$256. } We train our TEM-B model and evaluate it by providing object locations from real images and composing images according to the number of objects. We condition the energy network on a target object location $\mathbf{p}=(x,y)\in[0,1]^2$ (normalized image coordinates) through two complementary pathways: a global coordinate code injected by adaptive layer normalization, and a spatial bias added directly to the patch-token grid. To condition TEM on location, we encode each coordinate with a NeRF-style Fourier feature encoding,
\begin{equation}
    \gamma(c) = \big[c,\ \sin(2^0\pi c),\ \cos(2^0\pi c),\ \dots,\ \sin(2^{L-1}\pi c),\ \cos(2^{L-1}\pi c)\big],
    \label{eq:fourier}
\end{equation}
which gives a richer positional signal than raw coordinates~\citep{mildenhall2021nerf}. With $L{=}8$ frequency bands, $(x,y)$ maps to a $2(1{+}2L){=}34$-dimensional code, which is linearly projected, combined with the (optional) attribute embeddings, and injected into every transformer block via AdaLN. We train for 150k steps with batch size 64.

\textbf{Evaluation protocol.} Composable Diffusion~\citep{liu2022compositional}
operates at 128$\times$128, and its evaluation classifier is trained at this
resolution. We use their released 128$\times$128 CLEVR checkpoint directly, and
train the classifier ourselves following their training recipe with 128$\times$128 CLEVR dataset. TEM
generates at 256$\times$256, and we downsample its samples to 128$\times$128
only for scoring, so both methods are judged by the same classifier at the same
resolution. The classifier reaches over 99.9\% detection accuracy on
real 128$\times$128 CLEVR images, so the reported accuracies reflect generation
quality rather than detector error. We report detection success rate over 1{,}000
scenes for 1-, 2-, and 3-object compositions.

\section{Out-of-Distribution Detection}
\label{sec:appendix_ood}

\textbf{TEM as an OOD Detector.} TEM provides a scalar energy that we evaluate as an OOD score,
using lower energy to indicate more ID-like inputs. We
evaluate this property as a secondary capability of TEM. Specifically, we use a
TEM-XL model trained with class-label dropout: during training, the class
condition is randomly replaced with a null token, allowing the model to learn an
unconditional energy function $E_\theta(\mathbf{x})$ alongside the conditional
one. At test time, we query the unconditional branch and use the scalar energy
$S(\mathbf{x}) = -E_\theta(\mathbf{x})$ as the OOD score in a single forward
pass, with no class labels, post-hoc calibration, or OOD exposure.

\textbf{Diffusion Baselines.} As baselines we use pretrained DiT-XL/2~\citep{peebles2023scalable},
SiT-XL/2~\citep{ma2024sit}, and SiT-XL/2+REPA~\citep{yu2024representation} as
ImageNet-trained generative baselines. We use the official released checkpoints:
7M training iterations for DiT-XL/2 and SiT-XL/2, and 4M for SiT-XL/2+REPA. Each test image $x$ is encoded to a latent
$z_0 = \mathcal{E}(x)$ by the Stable Diffusion VAE. For DiT, we construct $z_{t} =
\sqrt{\bar{\alpha}_{t}}\,z_0 + \sqrt{1 -
\bar{\alpha}_{t}}\,\boldsymbol{\epsilon}$ with $\boldsymbol{\epsilon} \sim
\mathcal{N}(0, \mathbf{I})$, and the network is queried under the null class
token $y = \varnothing$ (the learned embedding at index 1000, used for
classifier-free guidance dropout during training). The OOD score is the
denoising residual
\begin{equation}
  s(x) = \left\| \boldsymbol{\epsilon} - \boldsymbol{\epsilon}_\theta(z_{t}, t, \varnothing) \right\|^2,
\end{equation}
where lower $s(x)$ indicates a more in-distribution input,
following~\citet{wyatt2022anoddpm}. We score DiT using its noise-prediction residual and
SiT-based models using their velocity-prediction residuals,
with each model's native interpolation and time convention. All baselines use a single network evaluation, matching the single-forward-pass
cost of TEM.  For DiT, we use timestep $t{=}350$. For SiT+REPA, we select the
timestep that matches the same noise level.

We include SiT-XL/2+REPA to separate two possible explanations for TEM's
performance. Since TEM is trained with PRA, its energy function inherits
semantic structure from VFM, and one might attribute the gain to that
supervision rather than to the energy score itself. SiT-XL/2+REPA receives the same DINOv2 alignment signal but is scored by its denoising residual, and its AUROC is close to that of SiT-XL/2 (Table~\ref{tab:ood}). Alignment to a vision
foundation model therefore does not by itself account for the gap. What
distinguishes TEM is that the energy provides a single scalar that directly
ranks inputs by fit to the learned distribution. 

\textbf{Evaluation Protocol.} Following~\citet{huang2021mos}, we use the ImageNet-1k validation set
(50{,}000 images) as the in-distribution reference and evaluate on four OOD
benchmarks spanning a range of semantic distances from ImageNet:
\textit{iNaturalist}~\citep{van2018inaturalist}, \textit{SUN}~\citep{xiao2010sun},
\textit{Places365}~\citep{zhou2017places}, and
\textit{Textures}~\citep{cimpoi2014describing}. We report AUROC ($\uparrow$).

\section{Additional related work}
\label{app:additionalrelated}
\subsection{Baselines in Table~\ref{tab:all_results}}
\label{app:baselines}

All baseline numbers are taken from the original papers; a dash indicates that the
corresponding setting was not reported. Among GANs we compare against BigGAN-deep
\citep{brock2018large} and GigaGAN \citep{kang2023scaling}. For autoregressive and masked
models we report MaskGIT \citep{chang2022maskgit} and VAR-d30 \citep{tian2024visual}. The
diffusion and flow baselines are ADM \citep{dhariwal2021diffusion}, LDM-4
\citep{rombach2022high}, DiT-XL/2 \citep{peebles2023scalable}, SiT-XL/2 \citep{ma2024sit},
and SiT-XL/2 with REPA \citep{yu2024representation}. Parameter counts refer to the
256$\times$256 configuration unless a resolution-specific model was released.
\subsection{Scaling Energy-based models}
A parallel line of work attains scale by learning energy landscapes from predictive targets rather than from negative samples drawn from the model itself. \citet{du2023reduce} assign a scalar energy to each diffusion timestep via denoising score matching, recovering MCMC-based composition but forgoing a single global energy. Equilibrium Matching~\citep{wang2025equilibrium} fits a time-invariant equilibrium gradient through a regression objective, defining the energy only implicitly through its gradient field. Its explicit scalar-energy variants (EqM-E) substantially degrade generation quality (FID 73.40 and 75.53 in the B/4 setting). Energy-Based Transformers~\citep{gladstone2025energy} learn an energy landscape by backpropagating a reconstruction loss through unrolled inner-loop optimization. Energy Matching~\citep{balcerak2025energy} explicitly parameterizes a
scalar potential, but achieves its high-fidelity generation results by
combining an optimal-transport objective with contrastive energy learning.
Its image generation experiments are limited to 32$\times$32 resolution
(CIFAR-10 and unconditional ImageNet 32$\times$32), so it is not included
in Table~\ref{tab:all_results}.

In contrast, TEM directly parameterizes a single scalar energy and trains
it contrastively using only negatives produced by its own MCMC sampler,
without a transport or flow-matching objective. Thus, while prior work
demonstrates that energy parameterizations can scale when coupled with
predictive, transport-based, or reconstruction-style objectives, our
results show that an explicit, contrastively trained EBM can itself scale
to ImageNet 256$\times$256.

\end{document}